\RequirePackage{fix-cm}
\pdfoutput=1
\documentclass[]{fairmeta}
\usepackage{amsmath,amsfonts,bm}

\def\eqref#1{equation~\ref{#1}}
\def\1{\bm{1}}

\DeclareMathAlphabet{\mathsfit}{\encodingdefault}{\sfdefault}{m}{sl}
\SetMathAlphabet{\mathsfit}{bold}{\encodingdefault}{\sfdefault}{bx}{n}

\newcommand{\E}{\mathbb{E}}

\newcommand{\R}{\mathbb{R}}

\newcommand{\reg}{\lambda}

\usepackage[utf8]{inputenc}
\usepackage[T1]{fontenc}
\usepackage{booktabs}
\usepackage{tabularx}
\usepackage{array}
\usepackage{multirow}
\usepackage{nicefrac}
\usepackage{microtype}
\usepackage[table]{xcolor}
\usepackage{colortbl}
\usepackage[normalem]{ulem}
\usepackage{graphicx}
\usepackage{subcaption}
\usepackage{wrapfig}
\usepackage{float}
\usepackage{tcolorbox}
\tcbuselibrary{breakable}
\usepackage{tikz}
\usepackage{pifont}
\usepackage{amssymb}
\usepackage{amsthm}
\usepackage{enumitem}
\usepackage{balance}
\usepackage{threeparttable}
\usepackage{placeins}

\newtheorem{theorem}{Theorem}
\newtheorem{lemma}[theorem]{Lemma}
\newtheorem{proposition}[theorem]{Proposition}

\theoremstyle{definition}

\newtheorem{assumption}{Assumption}

\newcommand{\dep}{\mathrm{dep}}
\newcommand{\Hreg}{\mathcal{H}_{\reg}}
\newcommand{\Hdep}{\mathcal{H}_{\dep}}
\newcommand{\bbeta}{\bm{\beta}}
\newcommand{\balpha}{\bm{\alpha}}
\newcommand{\bme}{\mathbf{e}}
\newcommand{\bmz}{\mathbf{z}}
\newcommand{\bmr}{\mathbf{r}}
\newcommand{\bmh}{\mathbf{h}}
\newcommand{\bmu}{\mathbf{u}}

\newcommand{\bmg}{\mathbf{g}}

\newcommand{\bmphi}{\bm{\phi}}
\newcommand{\bmR}{\mathbf{R}}

\newcommand{\hatz}{\widehat{\bmz}}
\newcommand{\hate}{\widehat{\bme}}
\newcommand{\Pperp}[1]{\mathbf{\Pi}^{\perp}_{#1}}
\newcommand{\spn}{\mathrm{span}}
\newcommand{\nablab}{\nabla_{\bbeta}}

\newcommand{\Sph}{\mathbb{S}}
\newcommand{\Emb}{\mathrm{Emb}}
\newcommand{\score}{S}
\newcommand{\bcP}{\mathcal{P}}

\newcommand{\bV}{\mathbb{V}}
\newcommand{\btheta}{\bm{\theta}}
\newcommand{\beeta}{\bm{\eta}}
\newcommand{\bTheta}{\bm{\Theta}}
\newcommand{\bPhi}{\bm{\Phi}}

\ifdefined\cleanrevision
  
  \DeclareRobustCommand{\old}[1]{}
\else
  
  \DeclareRobustCommand{\old}[1]{\sout{#1}\hspace{0.15em}}
\fi

\newcommand{\ms}[2]{\begin{tabular}[c]{@{}c@{}}#1\\[-2pt]\scriptsize$\pm$#2\end{tabular}}

\usepackage{pgfplots}
\usepgfplotslibrary{groupplots}
\usetikzlibrary{patterns}
\pgfplotsset{compat=1.17}
\definecolor{depred}{HTML}{B83832}
\definecolor{deltagray}{HTML}{4D4D4D}
\definecolor{gridgray}{HTML}{E6E6E6}
\definecolor{resblue}{HTML}{2F5A92}

\usepackage{flushend}
\usepackage{adjustbox}

\newcommand{\rightans}[1]{\textcolor{green!50!black}{\textbf{#1}}}
\newcommand{\wrongans}[1]{\textcolor{red!70!black}{\textbf{#1}}}

\definecolor{boxbg}{RGB}{249, 250, 251}      % near-white background
\definecolor{rowalt}{RGB}{243, 244, 246}     % zebra stripe
\definecolor{ruleline}{RGB}{203, 213, 225}   % soft slate rules
\definecolor{headbg}{RGB}{51, 65, 85}        % charcoal header bar
\definecolor{headfg}{RGB}{255, 255, 255}
\definecolor{labelcol}{RGB}{71, 85, 105}     % slate-600 labels
\definecolor{bodytext}{RGB}{30, 41, 59}      % slate-900 body
\definecolor{badgebg}{RGB}{226, 232, 240}    % token badge bg
\definecolor{badgefg}{RGB}{51, 65, 85}       % token badge fg
\definecolor{correct}{RGB}{29, 78, 216}      % blue-700
\definecolor{wrong}{RGB}{185, 28, 28}        % red-700

\newcommand{\tokbadge}[1]{%
  \tikz[baseline=(b.base)]\node[
    fill=badgebg, rounded corners=2pt,
    inner xsep=5pt, inner ysep=1.5pt,
    font=\bfseries\sffamily\scriptsize\color{badgefg}
  ](b){#1 tokens};\,%
}

\newcommand{\qualtitle}[1]{%
  \multicolumn{1}{@{}l@{}}{%
    \colorbox{headbg}{%
      \makebox[\dimexpr\linewidth-2\fboxsep\relax][l]{%
        \rule[-3pt]{0pt}{14pt}\hspace{6pt}%
        \textbf{\sffamily\color{headfg}~#1}%
      }%
    }%
  }
}

\newcommand{\qualcell}[3]{%
  \begin{minipage}[t]{\linewidth}
    \centering
    \fcolorbox{ruleline}{boxbg}{%
      \includegraphics[width=0.7\linewidth]{#1}%
    }\par
    \vspace{4pt}
    \raggedright\sffamily
    \textbf{\textcolor{labelcol}{Prompt:}}\ #2 \par
    \vspace{2pt}
    \textbf{\textcolor{labelcol}{Answer:}}\ \texttt{#3}
  \end{minipage}%
}

\newcolumntype{L}{>{\raggedright\arraybackslash}p{0.47\textwidth}}
\arrayrulecolor{ruleline}
\usepackage{xspace}
\def\eg{\emph{e.g.}\xspace}
\def\ie{\emph{i.e.}\xspace}

\title{Depth-Linked Residual Tokens for Multimodal \\ Retrieval and Generation}

\author[]{Jiangfan Zhang}
\author[]{Wentao Bao}
\author[]{Xian Wu}
\author[]{Xiangjun Fan}
\author[]{Jianpeng Cheng}
\author[]{Jun Xiao}

\affiliation[]{Meta AI}

\abstract{
Iterative architectures such as looped language models improve parameter efficiency and reasoning capacity of transformers, yet their sequential backbone iterations introduce substantial computational latency. Conversely, parallelizable learnable special tokens (such as the register tokens in Vision Transformer) enable efficient single-pass inference, but often suffer from the absence of sequential refinement and exhibit redundant representations across tokens. To address these challenges, we introduce \textbf{Depth}-L\textbf{ink}ed Residual Tokens (\textbf{DepThink}), an efficient framework designed for multimodal retrieval and generation. Operating within a single pre-filling forward pass, DepThink appends a sequence of K learnable residual tokens whose hidden states are coupled through a lightweight, learnable residual recurrence. This mechanism enables each residual token to progressively capture complementary information missed by preceding tokens. Theoretically, we prove that DepThink provides strictly expanded geometric directional reachability over standard learnable tokens and guarantees a provable one-step descent advantage. We evaluate DepThink across multimodal retrieval and generation tasks. DepThink consistently outperforms  learnable-token baselines and matches or exceeds looped architectures across dense (2B, 8B) and sparse mixture-of-experts (30B-A3B) backbones under both LoRA and full-parameter fine-tuning, while accelerating training by up to 11.9 and inference by 4.3 over looped architectures.
}

\date{\today}
\begin{document}

\maketitle

\section{Introduction}
\label{section:intro}

The evolution of modern foundation models has increasingly shifted from simply expanding parameter counts to maximizing the computational utility of existing weights. Among the most promising developments in this direction are looped architectures~\citep{yang2023looped,xu2024looped,saunshi2025looped}. By iteratively recycling representations through shared transformer layers, looped models achieve notable parameter efficiency and improved reasoning capacity, effectively allowing a model to ``think longer'' without inflating its parameter footprint. However, iterative approaches that loop entire transformer layers~\citep{hao2024training,zhu2025scaling} incur prohibitive computational latency.

An appealing alternative to sequential looping is augmenting the input with parallelizable, learnable special tokens---such as register tokens in Vision Transformers~\citep{darcet2023vision}, soft prompts~\citep{lester2021power,li2021prefix}, or visual query embeddings~\citep{li2023blip}. In non-autoregressive latent reasoning~\citep{goyal2310think,su2025token}, these tokens interact with the input context within a single, highly parallelizable forward pass. Nevertheless, standard unlinked tokens lack the iterative refinement mechanisms that give looped architectures their strength, leading to less expressive representations.

To resolve the trade-off between effeciecny and expressiveness, we propose \textbf{Dep}th-L\textbf{ink}ed Residual Tokens (\textbf{DepThink}), a lightweight framework that bridges parallel token computation and iterative refinement for multimodal retrieval and generation. Rather than executing recurrent passes through the entire backbone, DepThink appends a sequence of $K$ learnable tokens to the input and couples their contextualized representations via a learnable, first-order linear residual recurrence during the single pre-filling stage. By accumulating information through explicit depth-linked residual updates, each token progressively captures complementary features missed by earlier tokens. This architecture naturally serves dual purposes:
\begin{itemize}
    \item \textbf{Multimodal Retrieval:} It extracts multiple latent representations for  cross-modal retrieval, overcoming representation redundancy and expanding feature expressiveness.
    \item \textbf{Latent Deduction and Generation:} It augment prompt  latent states into an ordered deductive trajectory, simulating the stepwise deduction of looped models without autoregressive latency.
\end{itemize}

Beyond the analytical computational efficiency gains over looped architectures, we establish theoretical foundations for DepThink. We prove that DepThink subsumes standard  learnable token architectures and expands the per-input directional reachability.
We further prove that this geometric expansion translates into a one-step first-order descent advantage under gradient descent, ensuring  superior risk reduction.

Our contributions are summarized as follows:
\begin{itemize}
    \item We introduce Depth-Linked Residual Tokens (DepThink), a parameter-efficient token mechanism that emulates the iterative refinement of looped architectures within a single, parallel forward pass.
    \item We provide rigorous theoretical proofs establishing that DepThink achieves strictly greater directional reachability and a provable one-step gradient descent advantage over baseline with vanilla unlinked learnable tokens.
    \item We evaluate DepThink across diverse benchmarks spanning multimodal retrieval (MMEB-V2~\citep{meng2025vlm2vec}), math reasoning (GSM8K~\citep{cobbe2021training}, MATH~\citep{hendrycksmath2021}, and knowledge QA (NQ~\citep{kwiatkowski_natural}, HotpotQA~\citep{yang2018hotpotqa}).
    \item Extensive controlled experiments show that DepThink consistently outperforms vanilla unlinked \emph{learnable token baselines} and matches or exceeds \emph{looped architecture} across dense (2B, 8B) and sparse mixture-of-experts (30B-A3B) backbones under both LoRA and full-parameter tuning, while delivering up to $11.9\times$ faster training and $4.3\times$ faster inference over  looped architectures.
\end{itemize}

\section{Related Work}
\label{sec:related_work}

\textbf{Looped Architectures.}
Reusing parameters across network depth is a long-standing strategy for improving parameter efficiency. The Universal Transformer~\citep{dehghani2019universal} introduced this paradigm for sequence models, which has recently regained momentum for reducing memory footprints in edge deployment~\citep{li2025megrez2}, enabling test-time compute scaling~\citep{wu2025parallel,geiping2025recurrent}, and advancing latent reasoning~\citep{zhu2025scaling}. \citet{csordas2021neural} generalized Universal Transformers to mixture-of-experts architectures, \citet{saunshi2025looped} demonstrated their reasoning efficacy, and \citet{fan2024looped} established their length generalization capabilities. Further, \citet{prairie2026parcae} investigated scaling laws for looped structures, while \citet{zeitoun2026hyperloop} combined hyper-connections with looped models to enhance representation capacity. Despite these advantages, looped architectures are inherently constrained by sequential inference latency.

\textbf{Learnable Tokens.}
Augmenting transformer inputs with learnable  tokens is widely applied across NLP and vision. 
In language modeling, soft prompt tuning~\citep{lester2021power} and prefix tuning~\citep{li2021prefix} optimize task-specific virtual tokens while keeping base model weights frozen, whereas pause tokens~\citep{goyal2310think} append copies of a single learnable token to let the LLM perform extra computation before answering.
In vision transformers, register tokens~\citep{dosovitskiy2020image,darcet2023vision} eliminate feature artifacts and aggregate global context. In multimodal domains, prompt-tuning approaches such as CoOp~\citep{zhou2022learning} and CoCoOp~\citep{zhou2022conditional} adapt vision-language representations, whereas query-based architectures like BLIP-2's Q-Former~\citep{li2023blip}, InstructBLIP~\citep{dai2023instructblip}, and Flamingo's Perceiver Resampler~\citep{alayrac2022flamingo} use learnable queries as modality bridges. In retrieval, TTE-Flash~\citep{cheng2026tte} adopts learnable tokens for multi-vector retrieval. Compared to looped architectures, learnable tokens are parallelizable for efficient inference but lose the expressiveness of iterative state update. 

\textbf{Learnable Residual Connections.}
Parametric residual connections are historically employed to stabilize and accelerate optimization in very deep networks. 
Gates in Highway Networks~\citep{srivastava2015highway}, zero-initialized gates in ReZero~\citep{bachlechner2021rezero}, and channel-wise multipliers in LayerScale~\citep{touvron2021going} stabilize deep networks by scaling the layer-wise residual stream. DepThink uses a related algebraic form for a fundamentally different purpose: it couples distinct token positions along the sequence within one pre-filling pass, so that each token captures information missed by its predecessors.

\section{Methodology}
\label{sec:method}

\subsection{Preliminaries: Looped Architectures}
\label{sec:prelim}

Looped transformers, also referred to as LoopLMs~\citep{pmlr-v202-giannou23a,saunshi2025looped}, decouple effective computational depth from parameter count by recurrently applying a weight-tied causal transformer backbone across multiple iterations.

Let $T_{\btheta}$ be a causal transformer backbone with parameters $\btheta$, and let $\Emb(x) \in \R^{L \times D}$ be
the $L$ input embeddings of a multimodal prompt $x$. A looped transformer
(LoopLM)~\citep{pmlr-v202-giannou23a,saunshi2025looped} applies the weight-tied backbone $K$ times, $\bmh_{1:L}^{(k)}
\triangleq T_{\btheta}(\bmh_{1:L}^{(k-1)})$ with $\bmh_{1:L}^{(0)} \triangleq \Emb(x)$, recomputing all $L$ positions at
every step. A closely related instance of LoopLMs is \textbf{autoregressive latent reasoning}~\citep{hao2024training}, 
% which avoids re-encoding the input prompt by iteratively feeding the last token's output hidden state back into $T_{\btheta}$ as the next continuous input embedding for $K$ steps,
which avoids re-encoding the input prompt by iteratively feeding the last token's output hidden state back as the next input embedding, $\tilde{\bmz}_k(x) \triangleq [T_{\btheta}([\Emb(x); 
\tilde{\bmz}_{1:k-1}(x)])]_{L+k-1}$, so its $K$ latent steps remain sequential. DepThink removes this bottleneck: it replaces the sequential inputs $\tilde{\bmz}_{1:K-1}(x)$ with $K$ learnable tokens processed in a single pre-filling pass, and recovers cross-step refinement through a lightweight learnable residual recurrence.

\subsection{Depth-Linked Residual Tokens}
\label{sec:arch}

For any input $x$ (representing text prompts, images, or interleaved multimodal queries) with input embeddings $\Emb(x) \in \R^{L \times D}$, rather than executing sequential backbone iterations, we append a sequence of $K$ learnable token embeddings $\bmR = [\bmr_1; \dots; \bmr_K] \in \R^{K \times D}$ directly to the input sequence during the pre-filling stage. The unified forward pass is expressed as $[\bmh_{1:L}(x); \bme_{1:K}(x)] \triangleq
T_{\btheta}([\Emb(x); \bmr_{1:K}])$,
where $\bmh_{1:L}(x) \in \R^{L \times D}$ corresponds to the hidden states of the input sequence, and $\bme_{1:K}(x) \in \R^{K \times D}$ denotes the extracted hidden states of the $K$ appended tokens. Under causal attention masking, each token hidden state $\bme_k(x)$ attends to all input tokens $\Emb(x)$ as well as the prefix tokens $\bmr_{\le k}$.

\begin{figure}[t]
    \centering
    \includegraphics[width=0.9\linewidth]{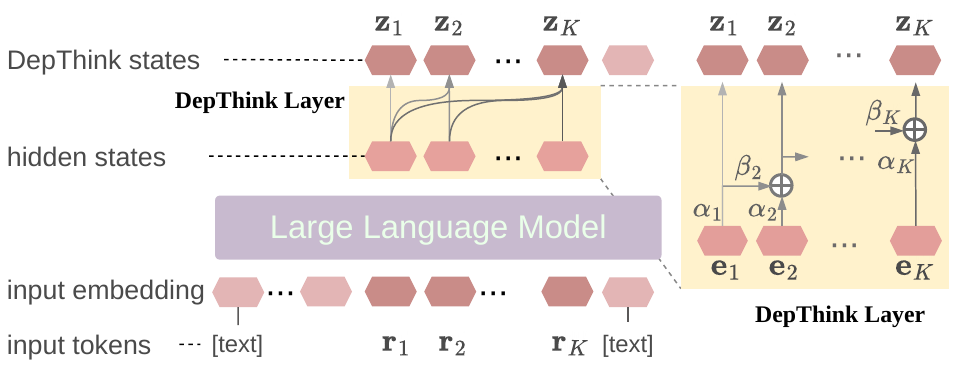}
    \caption{
    \textbf{Overview of Depth-Linked Residual Tokens (DepThink).}
    Given the input sequence $\Emb(x)$, DepThink appends $K$ learnable tokens $\{\bmr_1, \dots, \bmr_K\}$ in a single forward pass. Their contextualized hidden states $\{\bme_1, \dots, \bme_K\}$ are linked via a learnable first-order residual recurrence with parameters $\balpha$ and $\bbeta$, synthesizing $K$ progressively refined latent representations $\{\bmz_1, \dots, \bmz_K\}$ without autoregressive backbone passes.
    }
    \label{fig:depthink}
\end{figure}

\textbf{Learnable Residual Recurrence.}
As illustrated in Fig.~\ref{fig:depthink}, standard learnable tokens interact solely through self-attention. DepThink instead couples the token hidden states through a learnable, first-order linear recurrence.

Let $\balpha \triangleq [\alpha_1, \dots, \alpha_K] \in \R^K$ and $\bbeta \triangleq [\beta_1, \dots, \beta_K] \in \R^K$ denote learnable scaling and recurrence coefficients, respectively. The refined output of the $k$-th token, designated as a \textbf{residual latent representation}, is defined recursively as
\begin{equation}
\bmz_k(x) \;\triangleq\; \beta_k\, \bmz_{k-1}(x) \;+\; \alpha_k\, \bme_k(x)
\label{eq:iir_def}
\end{equation}
Unrolling~(\ref{eq:iir_def}) yields the closed-form accumulation $\bmz_k(x) = \sum_{i=1}^k M_{ki}(\balpha, \bbeta) \bme_i(x)$ 
where $M_{ki}(\balpha, \bbeta) \triangleq \alpha_i \prod_{j=i+1}^k \beta_j$, 
and we adopt the convention that the empty product equals $1$ when $i = k$ (hence $M_{kk} = \alpha_k$). 
The complete DepThink head is therefore
\begin{equation}
\bPhi^\dep(x) \triangleq \bigl[\bmz_1(x);\; \dots;\; \bmz_K(x)\bigr] \in \R^{K \times D}.
\label{eq:depth-linked_heads}
\end{equation}
By initializing $\balpha = \mathbf{1}$ and $\bbeta = \mathbf{0}$, DepThink smoothly recovers the vanilla unlinked \emph{learnable token baseline} $\bPhi^\reg(x) \triangleq \bme_{1:K}(x)$ at initialization, allowing stable optimization.

\subsection{Training for Retrieval or Generation}
\label{sec:objectives}

DepThink yields an ordered sequence of refined DepThink state vectors $\{\bmz_k\}_{k=1}^K$ that naturally unify multimodal representation learning and conditional generation. Training therefore depends completely on the task supervision.

\textbf{Multimodal Retrieval.}
In multimodal retrieval tasks, models require compact yet expressive multi-aspect representations. While unlinked learnable tokens may collapse into redundant subspaces, DepThink's recurrence can force token $k$ to capture complementary information missed by tokens $1, \dots, k-1$. We score a query $x$ against a target $y$ by the average per-token cosine similarity $S(x, y) \triangleq \frac{1}{K} \sum_{k=1}^{K} \bigl\langle \hat{\bmz}_k(x), \hat{\bmz}_k(y)\bigr\rangle$ 
where $\hat{\bmz}_k \triangleq \bmz_k / \|\boldsymbol{z}_k\|_2$ (and $\hat{\bmz}_k
\triangleq \boldsymbol{0}$ if $\bmz_k = \boldsymbol{0}$). For the unlinked learnable token baseline, $\bmz_k$ is replaced by $\bme_k$.
Given a batch of $N$ positive multimodal pairs $\{(x_i, y_i)\}_{i=1}^N$ drawn i.i.d.\ from data distribution $\bcP$ and temperature $\tau > 0$, we minimize the InfoNCE loss~\citep{oord2018representation} ${L}_R  \triangleq -\frac{1}{N} \sum_{i=1}^N \log \frac{\exp\bigl(\score(x_i, y_i)/\tau\bigr)}{\sum_{j=1}^N \exp\bigl(\score(x_i, y_j)/\tau\bigr)}$.

\textbf{Conditional Generation.}
In conditional generation tasks and similar to looped architectures, DepThink augments transformer hidden states to compute a latent  trajectory of chained transformer states, $\boldsymbol{\Phi}_i \triangleq \boldsymbol{\Phi}^{\mathrm{dep}}(x_i)$, prior to decoding, enriching representations for reasoning-intensive tasks (\eg, math deduction and multi-hop QA). Given prompt $x_i$ and target sequence $u_i = (u_i^1, \dots, u_i^m)$, the decoder predicts tokens autoregressively conditioned on prior tokens and the latent trajectory $\boldsymbol{\Phi}_i$ with the negative log-likelihood auto-regressive loss ${L}_{\mathrm{AR}} \triangleq -\frac{1}{N} \sum_{i=1}^{N} \sum_{j=1}^{m_i} \log P\bigl(u_i^{j} \mid u_i^{<j},
x_i, \boldsymbol{\Phi}_i\bigr).$ 
For the unlinked learnable token baseline, $\boldsymbol{\Phi}_i$ is replaced by $\bme_{1:K}(x_i)$ in the equation above.

\section{Theoretical Analysis}
\label{Section_Theory}

We first compare DepThink with the \emph{unlinked learnable token baseline}, whose head is $\bPhi^\reg_{(\btheta, \bmR)}(x) \triangleq \bme_{1:K}(x)$ for an input $x$ in the input space $\mathcal{X}$. Let $\beeta \triangleq (\btheta, \bmR, \balpha, \bbeta)$ collect the DepThink parameters, with backbone parameters $\btheta \in \bTheta$, and let $\bPhi^\dep_{\beeta}$ be the DepThink head in~(\ref{eq:depth-linked_heads}), with $\bmz_0 \triangleq \bm{0}$ in~(\ref{eq:iir_def}). For either objective in Section~\ref{sec:objectives}, $L_{\bcP}(\bPhi)$ denotes its population risk under $\bcP$, \ie, $\E_{\bcP^N}[{L}_R]$ for retrieval and $\E_{\bcP^N}[{L}_{\mathrm{AR}}]$ for generation, with $\bcP$ over $(x_i, u_i)$, computed with the backbone $(\btheta, \bmR)$ that produces $\bPhi$. The DepThink and the unlinked learnable token baseline thus differ only in the head.

\subsection{Lower Risk Guarantee via Subsumption} \label{section_risk}
Let $\Hreg \triangleq \{\bPhi^\reg_{(\btheta, \bmR)}\}$ and $\Hdep \triangleq \{\bPhi^\dep_{\beeta}\}$ over $(\btheta, \bmR) \in \bTheta \times \R^{K \times D}$ and $(\balpha, \bbeta) \in \R^K \times \R^K$, let $R^*(\mathcal{H}, \bcP)$ be the infimum of $L_{\bcP}$ over the parameters of $\mathcal{H}$, and define the lift
\begin{equation}
\beeta_0(\btheta, \bmR) \;\triangleq\; (\btheta, \bmR, \bm{1}, \bm{0}).
\label{eq:lift}
\end{equation}

\begin{proposition}[Lower Risk via Subsumption]
\label{prop:lift-id}
For any $(\btheta, \bmR) \in \bTheta \times \R^{K \times D}$, $\bPhi^\dep_{\beeta_0(\btheta, \bmR)} = \bPhi^\reg_{(\btheta, \bmR)}$ as functions on $\mathcal{X}$, and hence $L_{\bcP}(\bPhi^\dep_{\beeta_0(\btheta, \bmR)}) = L_{\bcP}(\bPhi^\reg_{(\btheta, \bmR)})$ for both the InfoNCE and the autoregressive risk. Consequently, $\Hreg \subseteq \Hdep$ and $R^*(\Hdep, \bcP) \le R^*(\Hreg, \bcP)$ for any $\bcP$ and batch size $N$.
\end{proposition}

The proof is in Appendix \ref{proof_prop_lift}. Proposition~\ref{prop:lift-id} ensures that DepThink subsumes the unlinked learnable token baseline. By providing a distortion-free embedding ($\beeta_0$) of any baseline checkpoint, it guarantees that DepThink can attain any baseline-attainable risk without an architectural penalty.

\subsection{%Geometric Advantage: 
Enhanced Directional Reachability% of DepThink
} 
\label{sec:reach}

Fix the backbone parameters $(\btheta, \bmR)$ and an input $x \in \mathcal{X}$. Let $\bV_k(x) \triangleq \spn\{\bme_1(x), \dots, \bme_k(x)\} \subseteq \R^D$ with $d_k(x) \triangleq \dim \bV_k(x)$, let $\hat{\bme}_k \triangleq \bme_k / \|\bme_k\|_2$, let $\Sph^{D-1} \triangleq \{\bmu \in \R^D : \|\bmu\|_2 = 1\}$ be the unit sphere, and collect the depth-$k$ directions reachable by varying $(\balpha, \bbeta) \in \R^K \times \R^K$ in
\begin{equation} \notag
\mathcal{D}^\dep_k(x; \btheta, \bmR) \triangleq \bigl\{\hat{\bmz}_k(x; \balpha, \bbeta) : \|\bmz_k\|_2 > 0\bigr\}.
\end{equation}

\begin{theorem}[Directional Reachability]
\label{lem:reach}
For any backbone configuration $(\btheta, \bmR)$, any $k \in \{1, \dots, K\}$, and any input $x$ with $\bV_k(x) \neq \{\bm{0}\}$,
\begin{equation}
\mathcal{D}^\dep_k(x; \btheta, \bmR) \;=\; \bV_k(x) \cap \Sph^{D-1},
\end{equation}
and, without normalization, $\{\bmz_k(x; \balpha, \bbeta)\} = \bV_k(x)$. In contrast, the baseline reaches only $\hat{\bme}_k(x)$ and $\bme_k(x)$, respectively.
\end{theorem}

The proof is in Appendix~\ref{proof_lemma_reach}. Theorem~\ref{lem:reach} formalizes the representational advantage intrinsic to  DepThink: a strict expansion of per-input directional freedom. For every input, the recurrence turns the single direction $\hat{\bme}_k(x)$ of the baseline into the whole $(d_k(x)-1)$-dimensional unit sphere of $\bV_k(x)$. Any direction spanned by the token states $\bme_{1:k}(x)$ is attainable at depth $k$, and, for generation, where $\bPhi$ is not normalized, so is any vector of $\bV_k(x)$.

\subsection{%Optimization Advantage: 
Provable One-Step Descent Superiority}
\label{section_opt_adv}
We now demonstrate that the enhanced directional reachability of DepThink established in Theorem~\ref{lem:reach} translates directly into a provable optimization advantage over the unlinked learnable token baseline. Start from an arbitrary baseline checkpoint $(\btheta, \bmR)$, lift it to $\beeta_0$ by~(\ref{eq:lift}), and focus on retrieval, letting $L_{\bcP}$ be the InfoNCE risk. With $\bmg^\reg \triangleq \nabla L_{\bcP}(\bPhi^\reg)|_{(\btheta, \bmR)}$, $\bmg^\dep \triangleq \nabla L_{\bcP}(\bPhi^\dep)|_{\beeta_0}$, and $\bmg^\dep_{\bbeta} \triangleq \nablab L_{\bcP}(\bPhi^\dep)|_{\beeta_0}$, one step of size $\mu > 0$ gives $(\btheta', \bmR') \triangleq (\btheta, \bmR) - \mu \bmg^\reg$ and $\beeta' \triangleq \beeta_0 - \mu \bmg^\dep$. We compare the two models through
\begin{equation}
\Delta(\mu) \;\triangleq\; L_{\bcP}\bigl(\bPhi^\reg_{(\btheta', \bmR')}\bigr) - L_{\bcP}\bigl(\bPhi^\dep_{\beeta'}\bigr).
\label{eq:gap-def}
\end{equation}
At the lift, the gradient decouples (Lemma~\ref{lem:lift-grad-id} in Appendix~\ref{app:subsec_gap_expansion}): the $(\btheta, \bmR)$-block of $\bmg^\dep$ equals $\bmg^\reg$, and the $\balpha$-block vanishes because $\ell_2$ normalization is scale-invariant. The only new direction is $\bbeta$. By Theorem~\ref{thm:beta-grad} (Appendix~\ref{app:subsec_lift_grads}), $\partial \hat L_N / \partial \beta_k$ at $\beeta_0$ is a softmax-weighted correlation between the target's (resp.\ query's) $\hat{\bme}_k$ and the query's (resp.\ target's) $\Pperp{\hat{\bme}_k} \bme_{k-1} / \|\bme_k\|_2$, where $\Pperp{\hat{\bme}_k}$ projects onto the orthogonal complement of $\hat{\bme}_k$. The vector $\Pperp{\hat{\bme}_k} \bme_{k-1}$ is thus the part of the preceding token state $\bme_{k-1}$ that is missing from the direction of the current token state $\bme_k$.

\begin{lemma}[Descent Gap Expansion]
\label{lem:gap-expansion_main}
Suppose Assumption~\ref{ass:smooth_app} (Appendix~\ref{app:subsec_gap_expansion}) holds: the token hidden states are almost surely nonzero, $\nabla L_{\bcP}(\bPhi^\dep_{\beeta})$ is $L_{\mathrm{loc}}$-Lipschitz on the closed ball $U_0$ of radius $r_{U_0}$ around $\beeta_0$, and the gradient can be taken under the expectation. Then, for $0 < \mu \le \mu_{\max} \triangleq \min(1/L_{\mathrm{loc}}, r_{U_0}/\|\bmg^\dep\|, r_{U_0}/\|\bmg^\reg\|)$ (with $r_{U_0}/0 \triangleq \infty$), which keeps both updates in $U_0$,
\begin{equation}
\Delta(\mu) \;=\; \mu \|\bmg^\dep_{\bbeta}\|^2 + O(\mu^2),
\label{eq:gap-expansion_main}
\end{equation}
with $|O(\mu^2)| \le \frac{\mu^2}{2} L_{\mathrm{loc}} (\|\bmg^\reg\|^2 + \|\bmg^\dep\|^2)$.
\end{lemma}

\begin{theorem}[Weak and Strict Optimization Advantage]
\label{thm:1step}
Under the assumptions of Lemma~\ref{lem:gap-expansion_main}, define
\begin{equation}
\bar{\mu} \;\triangleq\; \frac{2\|\bmg^\dep_{\bbeta}\|^2}{L_{\mathrm{loc}} \bigl(\|\bmg^\reg\|^2 + \|\bmg^\dep\|^2\bigr)}.
\label{eq:mu-bar}
\end{equation}
Then, for every step size covered by the two cases below, $\Delta(\mu) \ge 0$ (the DepThink step is never worse than the baseline step), with equality if and only if $\bmg^\dep_{\bbeta} = \bm{0}$.
\begin{enumerate}[label=(\roman*), leftmargin=*, nosep]
\item \textbf{Weak advantage.} If $\bmg^\dep_{\bbeta} = \bm{0}$, then $\beeta' = \beeta_0(\btheta', \bmR')$ and $\Delta(\mu) = 0$ for every $\mu > 0$.
\item \textbf{Strict advantage.} If $\bmg^\dep_{\bbeta} \neq \bm{0}$, then $\Delta(\mu) > 0$ for every $\mu \in (0, \min(\bar{\mu}, \mu_{\max}))$.
\end{enumerate}
\end{theorem}

The proofs are in Appendix~\ref{app:sec5_3_proofs}. The gain $\mu \|\bmg^\dep_{\bbeta}\|^2$ is exactly the squared gradient along the coordinates that the baseline lacks: a step increases $\beta_k$, re-injecting $\bmz_{k-1}$ into $\bmz_k$, precisely when, on average, the information discarded between depths $k-1$ and $k$ separates the positive target better than the softmax-weighted negatives. 
Notably, the advantage persists even where the baseline has converged: at a stationary point of the baseline ($\bmg^\reg
= \bm{0}$) with $\bmg^\dep_{\bbeta} \neq \bm{0}$, a sufficiently small DepThink step can still  lower the risk,  so the stationary points of the baseline are not traps for DepThink.

\subsection{Computational Efficiency Analysis}
\label{sec:efficient_inference}

A major advantage of DepThink over looped models is its computational efficiency in training and inference, for both
retrieval and generation. For retrieval, DepThink inference is a single pre-filling pass over the input and the $K$ tokens, followed by the recurrence~(\ref{eq:iir_def}). For generation, this pass is followed by standard autoregressive decoding conditioned on $\bPhi^\dep(x)$. Let $C_b = \mathcal{O}(L D^2 + L^2 D)$ be the pre-filling cost of $L$ tokens with hidden dimension $D$. A LoopLM that loops all layers $K$ times costs $C_{\text{loop}} = K C_b$, whereas DepThink costs $C_{\text{dep}} = \mathcal{O}\bigl((L+K)D^2 + (L+K)^2 D\bigr)$. Hence $C_{\text{loop}}/C_{\text{dep}} \ge K/(1+K/L)^2$ (with matching constants), which implies that DepThink is cheaper than LoopLMs whenever $(1+K/L)^2 < K$ (\eg, $L > 4$ for $K = 4$), and nearly $K$ times cheaper when
$K \ll L$, as in our setting ($K \le 16$ with hundreds to thousands of input tokens).

%%%%%%%%%%%%%%%%%%%%%%%%%%%%%%%%%%%%%%%%%%%%%%%%%%%%%%%%%%%%%%%%%%
%%%%%%%%%%%%%%%%%%%%% Experiments %%%%%%%%%%%%%%%%%%%%%%%%%%%%%%%%
%%%%%%%%%%%%%%%%%%%%%%%%%%%%%%%%%%%%%%%%%%%%%%%%%%%%%%%%%%%%%%%%%%

\section{Empirical Study}
\label{sec:exp}

\begin{table*}[t]
\centering
\small
\setlength{\tabcolsep}{3pt} % Reduced spacing slightly for better fit
\caption{Results on the MMEB-V2 benchmark for models \textbf{within 3B}.
Best results are in \textbf{bold}. Task abbreviations: CLS (classification),
QA (question answering), RET (retrieval), GD (grounding), MRET
(moment retrieval), VDR (ViDoRe), VR (VisRAG), and OOD (out-of-domain). For DepThink, we report the multi-run mean $\pm$ standard deviation. 
}
\label{tab:mmeb_full}
\resizebox{\textwidth}{!}{% Scale table to fit the text width
\begin{tabular}{l *{15}{c} c}
\toprule
\multirow{2}{*}{\textbf{Model}} 
& \multicolumn{5}{c}{\textbf{Image}} & \multicolumn{5}{c}{\textbf{Video}} &
\multicolumn{5}{c}{\textbf{VisDoc}} & \multirow{2}{*}{\textbf{All}} \\
\cmidrule(lr){2-6} \cmidrule(lr){7-11} \cmidrule(lr){12-16}
& \textbf{CLS} & \textbf{QA} & \textbf{RET} & \textbf{GD} & \textbf{All} &
\textbf{CLS} & \textbf{QA} & \textbf{RET} & \textbf{MRET} & \textbf{All} &
\textbf{VDR1} & \textbf{VDR2} & \textbf{VR} & \textbf{OOD} & \textbf{All} & \\
\midrule
\# of Datasets & 10 & 10 & 12 & 4 & 36 & 5 & 5 & 5 & 3 & 18 &
10 & 4 & 6 & 4 & 24 & 78 \\
\midrule 
[0.5\baselineskip]
\rowcolor{gray!15} \multicolumn{17}{c}{{Used extra training data (for reference)}} \\
LamRA~\citep{liu2025lamra} &
59.2 & 26.5 & 70.0 & 62.7 & 54.1 & 39.3 & 42.6 & 24.3 & 34.6 & 35.2 &
22.0 & 11.5 & 37.4 & 21.0 & 23.9 & 40.4 \\
GME~\citep{zhang2024gme} & 
54.4 & 29.9 & 66.9 & 55.5 & 51.9 & 34.9 & 42.0 & 25.6 & 32.4 & 33.9 &
86.1 & 54.0 & 82.5 & 43.1 & 72.7 & 54.1 \\
EmbeddingGemma-2~\citep{vera2026embeddinggemma2} &
-- & -- & -- & -- & 57.3 & -- & -- & -- & -- & 50.7 &
-- & -- & -- & -- & -- & 59.0 \\
AuroLA-Omni~\citep{xu2026scaling} &
63.9 & 58.3 & 63.4 & 78.6 & 65.1 & 51.2 & {56.9} & 42.0 & 43.1 & 48.9 &
79.7 & 54.9 & 79.3 & 41.9 & 69.2 & 62.6 \\
TTE-Flash~\citep{cheng2026tte}  &
67.1 & 61.7 & 70.7 & 81.3 & 68.3 & 54.4 & 51.5 & 45.4 & {51.5} & 50.6 &
75.6 & 52.7 & 83.8 & 41.1 & 68.1 & 64.1 \\
Ops-MM-Embedding~\citep{opensearch2025opsmm} &
{68.1} & 65.1 & 69.2 & 80.8 & 69.0 & 53.6 & 55.6 & 41.7 & 33.7 & 47.6 &
76.4 & 53.2 & 77.6 & 65.1 & 71.0 & 64.7 \\
RzenEmbed~\citep{jian2025rzenembed} &
65.3 & 61.7 & {73.8} & 77.8 & 68.5 & 45.6 & 47.5 & 38.3 & 36.7 & 42.6 &
{87.0} & 57.6 & 85.4 & 67.1 & 78.4 & 65.6 \\
UEmbed~\citep{song2026uembed} &
59.2 & {67.9} & 68.8 & {85.6} & 67.8 & {56.3} & 52.8 & 45.8 & 41.8 & 50.0 &
82.3 & {59.9} & 85.6 & {67.9} & 77.0 & 66.5

\\[0.5\baselineskip]
\rowcolor{gray!15} \multicolumn{17}{c}{{No extra training data}} \\

VLM2Vec~\citep{jiang2024vlm2vec} &
58.7 & 49.3 & 65.0 & 72.9 & 59.7 & 33.4 & 30.5 & 20.6 & 33.0 & 29.0 &
49.8 & 13.5 & 51.8 & 33.5 & 41.6 & 47.0 \\
DUME~\citep{zhang2024gme} &
59.3 & 55.0 & 66.3 & 78.0 & 62.5 & 37.7 & 46.6 & 17.1 & 30.0 & 33.2 &
67.6 & 43.3 & 47.1 & 33.8 & 52.8 & 52.7 \\
VLM2Vec-V2~\citep{meng2025vlm2vec} &
62.9 & 56.3 & 69.5 & 77.3 & 64.9 & 39.3 & 34.3 & 28.8 & 38.5 & 34.9 &
75.5 & 44.9 & 79.4 & 39.4 & 65.4 & 58.0 \\
UME-R1~\citep{lan2025umer1} &
64.8 & 62.8 & 67.6 & 77.2 & 66.6 & 44.3 & 51.2 & 32.9 & 39.7 & 42.2 &
72.4 & 46.2 & 79.2 & 37.2 & 63.9 & 60.1 \\
TTE$s$~\citep{cui2025think} & 
\textbf{67.9} & \textbf{66.6} & 70.2 & 84.1 & \textbf{70.1} & 47.3 & 49.1 & 34.4 & 33.2 & 32.1 &
77.5 & \textbf{53.2} & 83.2 & 41.1 & 68.8 & 63.1 \\
\textbf{DepThink (Ours)}
& \ms{65.8}{0.6} & \ms{66.5}{0.5} & \ms{\textbf{70.8}}{0.5} & \ms{\textbf{85.1}}{0.6} & \ms{69.8}{0.3}
& \ms{\textbf{54.7}}{0.9} & \ms{\textbf{55.5}}{0.3} & \ms{\textbf{48.1}}{0.4} & \ms{\textbf{47.9}}{0.6} & \ms{\textbf{51.9}}{0.4}
& \ms{\textbf{78.6}}{0.7} & \ms{53.0}{1.1} & \ms{\textbf{87.6}}{0.1} & \ms{\textbf{42.0}}{0.3} & \ms{\textbf{70.5}}{0.3}
& \ms{\textbf{65.9}}{0.2} \\
\bottomrule
\end{tabular}
}
\end{table*}

\textbf{Dataset setup.} We evaluate DepThink on multimodal retrieval benchmark MMEB-v2~\citep{meng2025vlm2vec} and reasoning-intensive generation tasks including the math and knowledge question answering. For math, we use GSM8K~\citep{cobbe2021training} and MATH~\citep{hendrycksmath2021} datasets. For knowledge QA, we evaluate models on Natural Questions (NQ)~\citep{kwiatkowski_natural} and HotpotQA~\citep{yang2018hotpotqa} datasets. For multimodal retrieval, we additionally compare against existing approaches on the CIRR~\citep{Liu_2021_ICCV}, MSCOCO~\citep{lin2014microsoft}, MSVD~\citep{chen2011collecting}, and QVHighlights~\citep{lei2021detecting}, which evaluate cross-modality retrieval among text, image, and video. 

\textbf{Comparisons.} We compare Depthink with two models: a vanilla unlinked learnable token baseline (without residual) and the LoopLM. For the LoopLM, we loop the full backbone (all layers) $K$ times, such that comparisons implicitly use the same reasoning steps. Unless noted, all baseline backbones are at the scale within 3B. 

\textbf{Implementation details.} We implement the DepThink primarily on the Qwen3-VL-Instruct 2B by default, that supports both text-only and multi-modality tasks. We use LoRA adapter of rank 64 for different tasks . We initialize $\boldsymbol{\alpha}$ and $\boldsymbol{\beta}$ to 1 and 0, respectively, so that DepThink starts from the learnable special-token baseline. We use $K=4$  tokens for retrieval and $K=16$ for text generation by default, and also provide comprehensive analysis on token scaling effect. Unless stated otherwise, training uses AdamW with learning rate $2\times10^{-4}$, weight decay 0.1, a cosine schedule with 500 warm-up steps, 20,000 updates on 32 H100 GPUs, and an effective batch size of 1,024.

\begin{table*}[t]
\centering
\footnotesize
\caption{Results of representative multi-modal retrieval tasks, where the modalities of retrieval pairs (``\emph{query}$\to$\emph{target}'') span over the text ($t$), image ($i$), and video ($v$). We include the LoopLM and DepThink without learnable residuals as controlled comparisons. For DepThink, we report the multi-run mean $\pm$ standard deviation in the last row.}
\label{tab:mmeb}
\setlength{\tabcolsep}{4pt}
\renewcommand{\arraystretch}{1.05}
\begin{tabular}[t]{lc}
\multicolumn{2}{c}{\textbf{CIRR} ($i+t\to i+t$)} \\
\toprule
Method & Recall \\
\midrule
CLIP~\citep{baldrati2022conditioned} & 42.8 \\
BLIP~\citep{liu2023candidate} & 44.8 \\
VLM2Vec-7B~\citep{jiang2024vlm2vec} & 51.1 \\
HINT~\citep{zhang2026hint} & 52.3 \\
UME-R1-7B~\citep{lan2025umer1} & 55.3 \\
Compodiff~\citep{gu2023compodiff} & 64.5 \\
\rowcolor{gray!15} LoopLM (2B) & 60.9 \\
\rowcolor{gray!15} DepThink w/o residual (2B) & 62.7 \\
\rowcolor{gray!15} \textbf{DepThink (2B)} & \textbf{65.8}{\tiny$\pm$1.1} \\
\bottomrule
\end{tabular}
\hspace{3em}
\begin{tabular}[t]{lc}
\multicolumn{2}{c}{\textbf{MSCOCO T2I} ($t\to i+t$)} \\
\toprule
Method & Recall \\
\midrule
CLIP~\citep{radford2021learning} & 37.8 \\
CoCa~\citep{yu2022coca} & 51.2 \\
BEIT-3~\citep{wang2023image} & 66.0 \\
EVA-CLIP~\citep{sun2024eva} & 73.6 \\
VLM2Vec-7B~\citep{jiang2024vlm2vec} & 77.2 \\
UME-R1-7B~\citep{lan2025umer1} & 78.3 \\
\rowcolor{gray!15} LoopLM (2B) & 78.2 \\
\rowcolor{gray!15} DepThink w/o residual (2B) & 78.6 \\
\rowcolor{gray!15} \textbf{DepThink (2B)} & \textbf{79.6}{\tiny$\pm$0.8} \\
\bottomrule
\end{tabular}

\vspace{8pt}

\begin{tabular}[t]{lc}
\multicolumn{2}{c}{\textbf{MSVD} ($t\to v+t$)} \\
\toprule
Method & Recall \\
\midrule
CLIP4Clip~\citep{luo2021clip4clip} & 45.2 \\
VLM2Vec-7B~\citep{jiang2024vlm2vec} & 46.7 \\
X-CLIP-B16~\citep{ma2022xclip} & 50.4 \\
UMT~\citep{li2023unmasked} & 58.2 \\
UME-R1-7B~\citep{lan2025umer1} & 60.8 \\
InternVideo2~\citep{wang2024internvideo2} & 61.4 \\
\rowcolor{gray!15} LoopLM (2B) & 74.9 \\
\rowcolor{gray!15} DepThink w/o residual (2B) & 71.5 \\
\rowcolor{gray!15} \textbf{DepThink (2B)} & \textbf{77.9}{\tiny$\pm$0.8} \\
\bottomrule
\end{tabular}
\hspace{3em}
\begin{tabular}[t]{lc}
\multicolumn{2}{c}{\textbf{QVHighlights} ($v+t\to v+t$)} \\
\toprule
Method & Recall \\
\midrule
UMT~\citep{li2023unmasked} & 57.7 \\
UME-R1-7B~\citep{lan2025umer1} & 54.9 \\
CLIP~\citep{baldrati2022conditioned} & 61.0 \\
M-DETR~\citep{lei2021detecting} & 60.2 \\
TaskWeave~\citep{yang2024task} & 64.3 \\
UVCOM~\citep{xiao2024bridging} & 64.5 \\
\rowcolor{gray!15} LoopLM (2B) & 67.7 \\
\rowcolor{gray!15} DepThink w/o residual (2B) & 65.1 \\
\rowcolor{gray!15} \textbf{DepThink (2B)} & \textbf{68.1}{\tiny$\pm$0.3} \\
\bottomrule
\end{tabular}
\end{table*}

\subsection{Main Results}

\textbf{Multimodal retrieval.} Tab.~\ref{tab:mmeb_full} reports the performance of DepThink alongside recent models on MMEB-v2 leaderboard. The models that were trained with extra training data are listed for reference. The benchmark results clearly show the better performance of DepThink. Especially, DepThink gained $+9.7\%$ over UME-R1~\citep{lan2025umer1} on video tasks and $+1.7\%$ over TTE~\citep{cui2025think} on VisDoc tasks, while keeping competitive on image tasks.
In Tab.~\ref{tab:mmeb}, we further compare against the LoopLM and no residual baselines, and include recent methods for reference. DepThink achieves the best performance on CIRR, MSCOCO T2I, and MSVD, and remains competitive on QVHighlights. Moreover, the gains over LoopLM/no residual baseline are especially pronounced on CIRR (+4.6/+2.9) and MSVD (+3.6/+5.7), validating our model design. 

\textbf{Math reasoning.} We compare with recent latent reasoning approaches, Coconut~\citep{hao2024training}, Pause Token~\citep{goyal2310think} and Token Assorted~\citep{su2025token} on GSM8K dataset in Tab.~\ref{tab:gsm8k}, and compare with HRPO~\citep{yue2026hybrid} and CRISP~\citep{li2026fixing} on MATH-500 in Tab.~\ref{tab:math} following~\citep{yue2026hybrid}. For fair comparison, Coconut is re-implemented with $K=16$ total number of reasoning steps using the same backbone as ours. 
The results show that DepThink reaches to the best performance at 71.11\% on GSM8K and 82.40\% on MATH-500. The relatively poor performance of Coconut can be attributed to its direct answer with implicit reasoning, while others explicitly write out reasoning traces before reaching to the final answer. 

\begin{table*}[t]
\centering
\footnotesize
\setlength{\tabcolsep}{4pt}
\setlength{\abovecaptionskip}{0pt}
\setlength{\belowcaptionskip}{3pt}
\renewcommand{\arraystretch}{1.05}
\begin{threeparttable}[t]
\caption{GSM8K accuracy (\%). $\dagger$: our 2B reproduction.}
\label{tab:gsm8k}
\begin{tabular}{lc}
\toprule
Method & Accuracy \\
\midrule
Pause Token~\citep{goyal2310think} & 53.70 \\
Token Assorted~\citep{su2025token} & 48.70 \\
Coconut~\citep{hao2024training} & 34.10 \\
\midrule
Coconut$^\dagger$ & 37.73 \\
LoopLM & 70.88 \\
\textbf{DepThink} & \textbf{71.11} \\
\quad (-residual) & 68.52 \\
\bottomrule
\end{tabular}
\end{threeparttable}
\hspace{4em}
\begin{threeparttable}[t]
\caption{MATH-500 accuracy (\%). $\dagger$: our 2B reproduction.}
\label{tab:math}
\begin{tabular}{lc}
\toprule
Method & Accuracy \\
\midrule
HRPO~\citep{yue2026hybrid} & 63.30 \\
CRISP~\citep{li2026fixing} & 76.00 \\
\midrule
Coconut$^\dagger$ & 71.00 \\
LoopLM & 79.80 \\
\textbf{DepThink} & \textbf{82.40} \\
\quad (-residual) & 81.60 \\
\bottomrule
\end{tabular}
\end{threeparttable}
\end{table*}

\begin{table*}[t]
\centering
\footnotesize
\setlength{\tabcolsep}{6pt}
\setlength{\abovecaptionskip}{0pt}
\setlength{\belowcaptionskip}{3pt}
\renewcommand{\arraystretch}{1.05}
\begin{threeparttable}
\caption{Exact Match (EM) on NQ and HotpotQA. Baselines are from Search-R1 and HRPO.}
\label{tab:nq_hotpotqa}
\begin{tabular}{lccc}
\toprule
\multirow{2}{*}{Method} & \multirow{2}{*}{NQ} & \multicolumn{2}{c}{HotpotQA} \\
\cmidrule(lr){3-4}
 & & w/o RAG & w/ RAG \\
\midrule
Qwen2.5-1.5B (sft)~\citep{yang2024qwen25} & 9.40 & 12.90 & -- \\
Qwen2.5-3B~\citep{yang2024qwen25} & 10.60 & 14.90 & -- \\
Search-o1~\citep{li2025searcho1} & -- & -- & 22.10 \\
Search-R1~\citep{jin2025searchr1} & -- & -- & 28.40 \\
HRPO (latent)~\citep{yue2026hybrid} & -- & -- & 31.60 \\
\midrule
LoopLM & 16.15 & 15.49 & 24.69 \\
\textbf{DepThink} & \textbf{16.57} & \textbf{15.81} & \textbf{31.81} \\
\quad (-residual) & 16.12 & 15.27 & 31.22 \\
\bottomrule
\end{tabular}
\end{threeparttable}
\end{table*}

\begin{figure}[t]
\centering
\begin{tikzpicture}[
    subcap/.style={font=\footnotesize, anchor=north, text width=5.5cm, align=center,
                   inner sep=1pt},
    gap/.style={font=\scriptsize\bfseries, text=deltagray},
]
\begin{groupplot}[
    group style={
        group size=2 by 1,
        horizontal sep=1.5cm,
    },
    width=0.38\columnwidth,
    height=3.6cm,
    scale only axis,
    % Axis & tick styling
    axis line style={draw=black!80, line width=0.6pt},
    tick style={draw=black!80, line width=0.5pt},
    major tick length=2pt,
    tick label style={font=\scriptsize, color=black!85, inner sep=1pt},
    label style={font=\footnotesize, color=black!90},
    % X-axis configuration (log2 scale: K doubles, so points are evenly spaced)
    xmode=log, log basis x=2,
    xlabel={number of special tokens},
    xlabel style={yshift=3pt},
    xmin=1.6, xmax=20,
    xtick={2, 4, 8, 16},
    xticklabels={2, 4, 8, 16},
    xminorticks=false,
    % Subtle grid
    grid=major,
    grid style={draw=gridgray, line width=0.45pt},
    clip=false,
    % Common plot styles
    every axis plot/.append style={
        line width=1.0pt,
        mark=square*,
        mark size=1.6pt,
    },
    % Per-panel legend (compact, boxed, short line samples)
    legend cell align=left,
    legend style={
        font=\scriptsize,
        draw=black!50,
        line width=0.4pt,
        fill=white,
        fill opacity=0.85,
        text opacity=1,
        inner sep=1pt,
        row sep=-1.5pt,
    },
    legend image code/.code={
        \draw[mark repeat=2, mark phase=2]
            plot coordinates {(0cm,0cm) (0.16cm,0cm) (0.32cm,0cm)};
    },
]

% =========================================================================
% Panel (a): Video retrieval
% =========================================================================
\nextgroupplot[
    ylabel={Video score (aggregated)},
    ylabel style={yshift=-2pt},
    ymin=0.50, ymax=0.53,
    ytick={0.50, 0.51, 0.52, 0.53},
    yticklabel style={
        /pgf/number format/fixed,
        /pgf/number format/precision=2,
        /pgf/number format/zerofill,
    },
    legend pos=south east,
]
    % DepThink
    \addplot[color=depred, mark options={fill=depred}] coordinates {
        (2,  0.5209)
        (4,  0.5225)
        (8,  0.5236)
        (16, 0.5290)
    };
    \addlegendentry{DepThink}

    \addplot[color=resblue, mark options={fill=resblue}] coordinates {
        (2,  0.5075)
        (4,  0.5123)
        (8,  0.5160)
        (16, 0.5207)
    };
    \addlegendentry{w/o residual}

    % Performance gain annotations (+pt)
    \node[gap] at (axis cs:2.22, 0.5142) {+1.34pt};
    \node[gap] at (axis cs:4,    0.5180) {+1.28pt};
    \node[gap] at (axis cs:8,    0.5205) {+0.76pt};
    \node[gap] at (axis cs:14.4, 0.5240) {+0.83pt};

% =========================================================================
% Panel (b): Math reasoning on GSM8K (epoch 5)
% =========================================================================
\nextgroupplot[
    ylabel={Math accuracy (\%)},
    ylabel style={yshift=-2pt},
    ymin=68, ymax=73,
    ytick={68, 69, 70, 71, 72, 73},
    legend pos=south east,
]
    % DepThink
    \addplot[color=depred, mark options={fill=depred}] coordinates {
        (2,  71.04)
        (4,  71.27)
        (8,  71.80)
        (16, 71.80)
    };
    \addlegendentry{DepThink}

    \addplot[color=resblue, mark options={fill=resblue}] coordinates {
        (2,  69.89)
        (4,  70.96)
        (8,  71.04)
        (16, 70.81)
    };
    \addlegendentry{w/o residual}

    % Performance gain annotations (+pt)
    \node[gap, anchor=north west] at (axis cs:1.75, 69.80) {+1.14pt};
    \node[gap, anchor=north west] at (axis cs:4.2,  70.90) {+0.30pt};
    \node[gap]                    at (axis cs:8.5,  71.45) {+0.76pt};
    \node[gap]                    at (axis cs:14.4, 71.15) {+0.99pt};

\end{groupplot}

% Subcaptions sit just below each panel's tick labels (outer south = incl. tick labels)
\node[subcap] at ([yshift=-1pt]group c1r1.outer south) {(a) MMEB video retrieval.};
\node[subcap] at ([yshift=-1pt]group c2r1.outer south) {(b) GSM8K math reasoning.};

\end{tikzpicture}
\setlength{\abovecaptionskip}{3pt}
\caption{Analysis of the number of learnable tokens for multimodal retrieval and math reasoning. ``w/o residual'' denotes DepThink without the residual connection.}
\label{fig:scaling}
\end{figure}
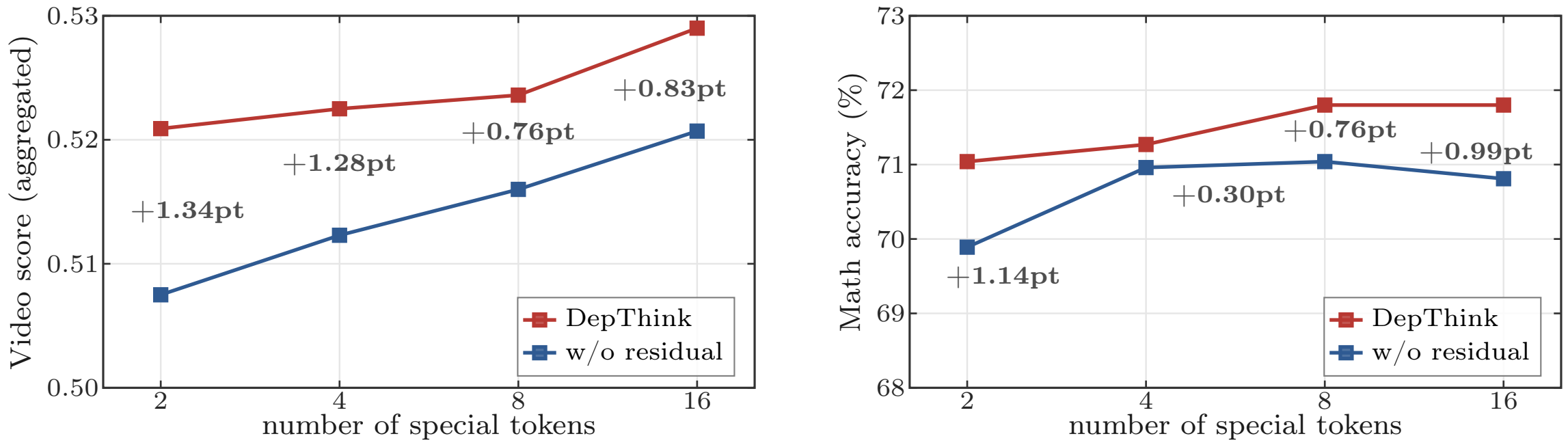

\textbf{Knowledge QA.} Beyond math, we evaluate DepThink on knowledge-intensive open-domain and multi-hop question answering. Following prior work~\citep{yue2026hybrid,jin2025searchr1}, we train models on the merged NQ~\citep{kwiatkowski_natural} and HotpotQA~\citep{yang2018hotpotqa} training splits and evaluate on their test splits by considering whether using retrieval-augmented generation (RAG), \ie, use top-10 retrieved documents as input context. As shown in Tab.~\ref{tab:nq_hotpotqa}, DepThink performs the best for both the with-and without RAG settings. In the RAG setting, LoopLM is inferior because its recycling of latents and hidden states from previous loop often disrupts the search query for context retrieval, whereas DepThink's special tokens are input-independent, so its search queries stay intact.

\begin{figure}[t]
\centering
\pgfplotsset{compat=1.17}
% =========================================================================
% Fig. 3: model scaling, averaged retrieval score
% =========================================================================
\begin{minipage}[t]{0.49\columnwidth}
\centering
\begin{tikzpicture}[
    baseline=(scaleax.south),
    gap/.style={font=\scriptsize\bfseries, text=deltagray, anchor=south, yshift=1pt},
    barloop/.style={/pgf/bar width=0.25, /pgf/bar shift=-0.27,
                    draw=black!80, line width=0.4pt, fill=resblue},
    bardep/.style={/pgf/bar width=0.25, /pgf/bar shift=0,
                   draw=black!80, line width=0.4pt, fill=depred},
    barnores/.style={/pgf/bar width=0.25, /pgf/bar shift=0.27,
                     draw=depred, line width=0.4pt, fill=white,
                     postaction={pattern=north east lines, pattern color=depred}},
]
\begin{axis}[
    name=scaleax,
    width=0.78\linewidth,
    height=3.6cm,
    scale only axis,
    % Axis & tick styling
    axis line style={draw=black!80, line width=0.6pt},
    tick style={draw=black!80, line width=0.5pt},
    major tick length=2pt,
    tick label style={font=\scriptsize, color=black!85, inner sep=1pt},
    label style={font=\footnotesize, color=black!90},
    ybar,
    xmin=-0.55, xmax=2.55,
    xtick={0, 1, 2},
    xticklabels={{2B\\full FT}, {8B\\LoRA}, {30B-A3B\\MoE, LoRA}},
    xticklabel style={align=center},
    x tick style={draw=none},
    ylabel={Avg.\ retrieval score},
    ylabel style={yshift=-2pt},
    % Headroom above the bars and gain labels for the legend
    ymin=65, ymax=88,
    ytick={65, 70, 75, 80, 85},
    % Subtle horizontal grid
    ymajorgrids=true,
    grid style={draw=gridgray, line width=0.45pt},
    % Bars start below ymin, so clip them; mode=individual keeps the gap labels unclipped
    clip mode=individual,
    % Small legend inside, top left
    legend cell align=left,
    legend style={
        font=\scriptsize,
        draw=black!50,
        line width=0.3pt,
        fill=white,
        inner sep=1pt,
        row sep=-1pt,
        at={(0.02, 0.98)},
        anchor=north west,
    },
    legend image code/.code={
        \draw[#1] (0cm, -0.08cm) rectangle (0.3cm, 0.1cm);
    },
]
    % Mean of CIRR, MSCOCO T2I, MSVD, QVHighlights
    \addplot[barloop] coordinates {(0, 71.105) (1, 74.7225) (2, 75.4575)};
    \addlegendentry{LoopLM}
    \addplot[bardep] coordinates {(0, 74.200) (1, 77.800) (2, 77.525)};
    \addlegendentry{DepThink}
    \addplot[barnores] coordinates {(0, 70.675) (1, 76.375) (2, 75.925)};
    \addlegendentry{w/o residual}

    % DepThink gain over LoopLM
    \node[gap] at (axis cs:0, 74.200) {+3.1};
    \node[gap] at (axis cs:1, 77.800) {+3.1};
    \node[gap] at (axis cs:2, 77.525) {+2.1};
\end{axis}
\end{tikzpicture}
\setlength{\abovecaptionskip}{3pt}
\caption{Retrieval score vs.\ model scale, averaged over four benchmarks. Gray: gain over LoopLM.}
\label{fig:retrieval_scaling}
\end{minipage}\hfill
% =========================================================================
% Fig. 4: wall time per step, training and inference
% =========================================================================
\begin{minipage}[t]{0.49\columnwidth}
\centering
\begin{tikzpicture}[
    baseline=(wallax.south),
    gap/.style={font=\scriptsize\bfseries, text=deltagray, align=center,
                inner sep=0.5pt, execute at begin node={\setlength{\baselineskip}{8pt}}},
]
\begin{axis}[
    name=wallax,
    width=0.78\linewidth,
    height=3.6cm,
    scale only axis,
    % Axis & tick styling
    axis line style={draw=black!80, line width=0.6pt},
    tick style={draw=black!80, line width=0.5pt},
    major tick length=2pt,
    tick label style={font=\scriptsize, color=black!85, inner sep=1pt},
    label style={font=\footnotesize, color=black!90},
    % X-axis configuration (log2 scale: K doubles, so points are evenly spaced)
    xmode=log, log basis x=2,
    xlabel={$K$},
    xlabel style={yshift=3pt},
    xmin=1.4, xmax=23,
    xtick={2, 4, 8, 16},
    xticklabels={2, 4, 8, 16},
    xminorticks=false,
    ylabel={Wall time / step (s)},
    ylabel style={yshift=-2pt},
    ymin=0, ymax=25,
    ytick={0, 5, 10, 15, 20, 25},
    % Subtle grid
    grid=major,
    grid style={draw=gridgray, line width=0.45pt},
    clip=false,
    % Training = solid + filled squares; inference = dashed + open squares
    train/.style={line width=1.0pt, mark=square*, mark size=1.6pt,
                  color=#1, mark options={fill=#1, solid}},
    infer/.style={line width=1.0pt, dashed, mark=square*, mark size=1.6pt,
                  color=#1, mark options={fill=white, solid}},
    % Small legend inside, top left (methods only; line style is in the caption)
    legend cell align=left,
    legend style={
        font=\scriptsize,
        draw=black!50,
        line width=0.3pt,
        fill=white,
        inner sep=1pt,
        row sep=-1pt,
        at={(0.02, 0.98)},
        anchor=north west,
    },
    legend image code/.code={
        \draw[mark repeat=2, mark phase=2]
            plot coordinates {(0cm,0cm) (0.16cm,0cm) (0.32cm,0cm)};
    },
]
    % --- LoopLM ---
    \addplot[train=resblue] coordinates {(2, 3.13) (4, 5.77) (8, 11.25) (16, 22.04)};
    \addlegendentry{LoopLM}
    \addplot[infer=resblue, forget plot] coordinates {(2, 2.01) (4, 2.73) (8, 4.20) (16, 7.24)};

    % --- DepThink ---
    \addplot[train=depred] coordinates {(2, 1.74) (4, 1.91) (8, 1.81) (16, 1.85)};
    \addlegendentry{DepThink}
    \addplot[infer=depred, forget plot] coordinates {(2, 1.69) (4, 1.71) (8, 1.68) (16, 1.70)};

    % --- w/o residual ---
    \addplot[train=black!55] coordinates {(2, 1.74) (4, 1.80) (8, 1.75) (16, 1.78)};
    \addlegendentry{w/o residual}
    \addplot[infer=black!55, forget plot] coordinates {(2, 1.66) (4, 1.80) (8, 1.69) (16, 1.67)};

    % DepThink speedup over LoopLM: train (top) / inference (bottom),
    % stacked on the solid LoopLM curve
    \node[gap, anchor=south, yshift=3pt]                   at (axis cs:2,  3.13)  {1.8$\times$\\1.2$\times$};
    \node[gap, anchor=south east, xshift=-1pt, yshift=1pt] at (axis cs:4,  5.77)  {3.0$\times$\\1.6$\times$};
    \node[gap, anchor=north west, xshift=1pt, yshift=-1pt] at (axis cs:8,  11.25) {6.2$\times$\\2.5$\times$};
    \node[gap, anchor=east, xshift=-3pt, yshift=3pt]       at (axis cs:16, 22.04) {11.9$\times$\\4.3$\times$};
\end{axis}
\end{tikzpicture}
\setlength{\abovecaptionskip}{3pt}
\caption{Wall time per step; solid: train, dashed: inference. Gray: speedup over LoopLM (train/inference).}
\label{fig:walltime}
\end{minipage}
\end{figure}

\subsection{Ablations and Analysis.}

\textbf{Token count scaling.} We evaluate the impact of varying the number of learnable tokens $K$ across multimodal retrieval and generation tasks in Fig.~\ref{fig:scaling}. On both GSM8K generation and MMEB video retrieval tasks, the performance of DepThink is improved as we scale up to 16 tokens. Meanwhile, on all token numbers, the performance of DepThink is above learnable token baselines.

\textbf{Model scaling.} In Fig.~\ref{fig:retrieval_scaling}, we scale the DepThink model size up to 8B and 30B-A3B with MoE, and also compare between LoRA and full fine-tuning (FT) recipes. These results showed that 8B and 30B-A3B models achieved much higher performance than 2B models for both tasks, while their performance differs by datasets. 
In Appendix~\ref{app:scaling_compare}, the detailed per-dataset comparisons on multimodal retrieval and GSM8K math reasoning are presented.

\textbf{Latency study.} In Fig.~\ref{fig:walltime}, we benchmark DepThink, DepThink without residual, and LoopLM models (see details in Appendix~\ref{app:efficiency}). It shows DepThink and no residual baseline remain almost constant latency as $K$ increases, showing that depth-linking itself adds little wall-clock overhead. LoopLM grows approximately linearly because it executes the backbone sequentially. At $K=16$, DepThink is $11.9\times$ faster per training step and $4.3\times$ faster in inference. 

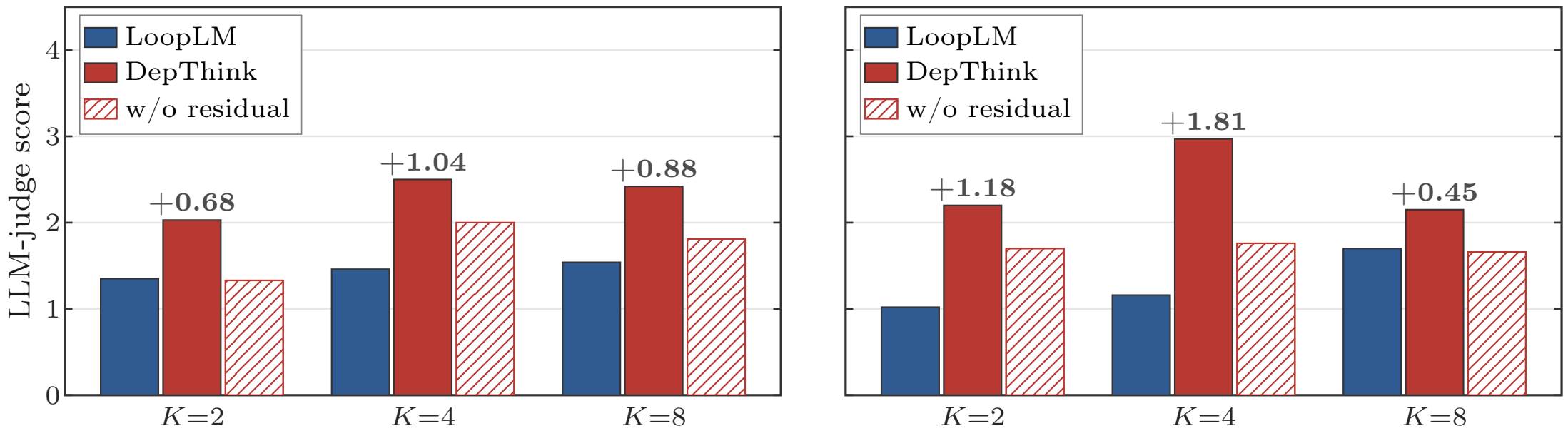
\begin{figure}[t]
\centering
\pgfplotsset{compat=1.17}
\begin{tikzpicture}[
    subcap/.style={font=\footnotesize, anchor=north, align=center, inner sep=1pt},
    gap/.style={font=\scriptsize\bfseries, text=deltagray,
                anchor=south, yshift=0.5pt, inner sep=0.5pt},
    barloop/.style={/pgf/bar width=0.25, /pgf/bar shift=-0.27,
                    draw=black!80, line width=0.4pt, fill=resblue},
    bardep/.style={/pgf/bar width=0.25, /pgf/bar shift=0,
                   draw=black!80, line width=0.4pt, fill=depred},
    barnores/.style={/pgf/bar width=0.25, /pgf/bar shift=0.27,
                     draw=depred, line width=0.4pt, fill=white,
                     postaction={pattern=north east lines, pattern color=depred}},
]
\begin{groupplot}[
    group style={
        group size=2 by 1,
        horizontal sep=0.6cm,
        y descriptions at=edge left,
    },
    width=0.40\columnwidth,
    height=3.6cm,
    scale only axis,
    % Axis & tick styling
    axis line style={draw=black!80, line width=0.6pt},
    tick style={draw=black!80, line width=0.5pt},
    major tick length=2pt,
    tick label style={font=\scriptsize, color=black!85, inner sep=1pt},
    label style={font=\footnotesize, color=black!90},
    ybar,
    xmin=-0.55, xmax=2.55,
    xtick={0, 1, 2},
    xticklabels={{$K{=}2$}, {$K{=}4$}, {$K{=}8$}},
    x tick style={draw=none},
    ymin=0, ymax=4.5,
    ytick={0, 1, 2, 3, 4},
    ylabel={LLM-judge score},
    ylabel style={yshift=-2pt},
    % Subtle horizontal grid
    ymajorgrids=true,
    grid style={draw=gridgray, line width=0.45pt},
    clip mode=individual,
    % Small legend inside each panel, top left
    legend cell align=left,
    legend style={
        font=\scriptsize,
        draw=black!50,
        line width=0.3pt,
        fill=white,
        inner sep=1pt,
        row sep=-1pt,
        at={(0.02, 0.98)},
        anchor=north west,
    },
    legend image code/.code={
        \draw[#1] (0cm, -0.08cm) rectangle (0.3cm, 0.1cm);
    },
]

% =========================================================================
% Panel (a): Semantics
% =========================================================================
\nextgroupplot
    \addplot[barloop]  coordinates {(0, 1.35) (1, 1.46) (2, 1.54)};
    \addlegendentry{LoopLM}
    \addplot[bardep]   coordinates {(0, 2.03) (1, 2.50) (2, 2.42)};
    \addlegendentry{DepThink}
    \addplot[barnores] coordinates {(0, 1.33) (1, 2.00) (2, 1.81)};
    \addlegendentry{w/o residual}

    % DepThink gain over LoopLM
    \node[gap] at (axis cs:0, 2.03) {+0.68};
    \node[gap] at (axis cs:1, 2.50) {+1.04};
    \node[gap] at (axis cs:2, 2.42) {+0.88};

% =========================================================================
% Panel (b): Coherence
% =========================================================================
\nextgroupplot
    \addplot[barloop]  coordinates {(0, 1.02) (1, 1.16) (2, 1.70)};
    \addlegendentry{LoopLM}
    \addplot[bardep]   coordinates {(0, 2.20) (1, 2.97) (2, 2.15)};
    \addlegendentry{DepThink}
    \addplot[barnores] coordinates {(0, 1.70) (1, 1.76) (2, 1.66)};
    \addlegendentry{w/o residual}

    \node[gap] at (axis cs:0, 2.20) {+1.18};
    \node[gap] at (axis cs:1, 2.97) {+1.81};
    \node[gap] at (axis cs:2, 2.15) {+0.45};

\end{groupplot}

\node[subcap] at ([yshift=-1pt]group c1r1.outer south) {(a) Semantics};
\node[subcap] at ([yshift=-1pt]group c2r1.outer south) {(b) Coherence};

\end{tikzpicture}
\setlength{\abovecaptionskip}{3pt}
\caption{Semantic and coherence evaluation of multimodal reasoning on OK-VQA via LLM-as-judge. Gray numbers give DepThink's gain over LoopLM.}
\label{fig:okvqa_judge}
\end{figure}

\textbf{Quality of multimodal reasoning.} To qualitatively show the of multimodal reasoning capability, we train models to generate CoT traces on OK-VQA~\citep{marino2019ok}, which contains high-quality CoT annotations. The Gemini-2.5-Pro is used as an LLM judge to score the semantic consistency and coherence (with the prompt in Appendix~\ref{sec:prompts-llm-judges}). As shown in Fig.~\ref{fig:okvqa_judge}, %Tab.~\ref{tab:llmjudge}, 
DepThink scores highest on both metrics, outperforming the no residual baseline and LoopLM. In the Appendix~\ref{sec:text-generation-examples}, demo examples are provided.

\section{Conclusion}
\label{sec:con}

We introduced DepThink, a non-autoregressive latent state expansion method with improve effeciecy over looped models. By linking special tokens across depth with gated residual connections, DepThink performs iterative refinement within a single pre-filling pass. Theoretically, we showed that DepThink attains greater directional reachability and a one-step descent advantage over standard learnable special-token baselines. Empirically, we demonstrate that DepThink excels at both multimodal retrieval and conditional generation tasks with significant efficiency gain over looped architecture. 

% {\small \textbf{LLM Usage Statement.} We used LLMs for experiment planning, basic coding, reporting results, and improving manuscripts' readability.
% Additionally, we used LLMs for grammar checking to improve the readability, format references, and presenting figures and tables in latex. We have reviewed all AI-assisted work such as the core coding, reporting numbers, hallucination checking for references, etc. We take responsibility for the final content of this work, including text, claims or artifacts produced with AI assistance.}

\FloatBarrier
\bibliographystyle{assets/plainnat}
\bibliography{paper}

\clearpage
\newpage
\beginappendix

% \section{First appendix}
% \clearpage
% \setcounter{page}{1}
% \maketitlesupplementary

\section{Proof of Proposition~\ref{prop:lift-id}} \label{proof_prop_lift}
% \begin{proof}
% By fixing the IIR parameters to $\balpha = \bm{1}$ and $\bbeta = \bm{0}$, the recurrence in (\ref{eq:iir_def}) simplifies to $\bmz_k = 0 \cdot \bmz_{k-1} + 1 \cdot \bme_k = \bme_k$ for all $k$. Thus,
% $\hatz_k = \hate_k$ wherever $\|\bme_k\|_2 > 0$, ensuring the heads of DepThink and Register yield identical
% multi-vector representations. This point-wise equivalence establishes the set inclusion $\Hreg \subseteq \Hdep$.
% The inequality regarding the optimal risks then follows directly from the fundamental property of infima:
% $\inf_{h \in \mathcal{H}_1} f(h) \ge \inf_{h \in \mathcal{H}_2} f(h)$ whenever $\mathcal{H}_1 \subseteq \mathcal{H}_2$.
% \end{proof}
\begin{proof}
At the lift $\beeta_0(\btheta, \bmR) = (\btheta, \bmR, \bm{1}, \bm{0})$, the recurrence in (\ref{eq:iir_def}) reduces to $\bmz_k(x) = 0 \cdot \bmz_{k-1}(x) + 1 \cdot \bme_k(x) = \bme_k(x)$ for every $k \in \{1, \dots, K\}$ and every input $x \in \mathcal{X}$. Hence $\bPhi^\dep_{\beeta_0(\btheta, \bmR)}(x) = \bme_{1:K}(x) = \bPhi^\reg_{(\btheta, \bmR)}(x)$ for all $x \in \mathcal{X}$, and in particular $\hatz_k(x) = \hate_k(x)$, including the convention $\hatz_k = \bm{0}$ when $\bme_k(x) = \bm{0}$.
Both risks depend on the model only through the head and the shared backbone $(\btheta, \bmR)$. The InfoNCE risk depends on the head through the scores $\score(x_i, y_j)$, which are computed from its normalized rows, and the autoregressive risk depends on it through the conditioning $\bPhi_i$. Therefore $L_{\bcP}(\bPhi^\dep_{\beeta_0(\btheta, \bmR)}) = L_{\bcP}(\bPhi^\reg_{(\btheta, \bmR)})$ for both objectives, any $\bcP$, and any batch size $N$.
Since every unlinked learnable token baseline head $\bPhi^\reg_{(\btheta, \bmR)}$ is itself the DepThink head $\bPhi^\dep_{\beeta_0(\btheta, \bmR)}$, we have $\Hreg \subseteq \Hdep$, and $R^*(\Hdep, \bcP) \le R^*(\Hreg, \bcP)$ follows because an infimum over a larger set cannot be larger.
\end{proof}

\section{Proof of Theorem \ref{lem:reach}} \label{proof_lemma_reach}
% \begin{proof}
% ($\subseteq$).
% By the closed-form expression in \eqref{eq:closed}, $\bmz_k(x; \balpha, \bbeta) = \sum_{i=1}^k M_{ki}(\balpha, \bbeta)\, \bme_i(x) \in \bV_k(x)$. Therefore, its normalized counterpart satisfies $\hatz_k = \bmz_k / \|\bmz_k\|_2 \in \bV_k(x) \cap \Sph^{D-1}$.

% ($\supseteq$).
% Let $\bmu \in \bV_k(x) \cap \Sph^{D-1}$. By the definition of $\bV_k$, there exist coefficients $c_1, \dots, c_k \in \R$ such that $\bmu = \sum_{i=1}^k c_i\, \bme_i(x)$. We construct the specific IIR coefficients:
% \begin{equation*}
% \begin{aligned}
% \balpha^* &\triangleq [c_1,c_2,\dots,c_k,0,\dots,0] \in \R^K,\\
% \bbeta^* &\triangleq [0,1,1,\dots,1] \in \R^K.
% \end{aligned}
% \end{equation*}
% Both rigorously lie within their respective parameter spaces. For $i \le k$, the product term simplifies to $\prod_{j=i+1}^k \beta^*_j = 1$, yielding $M_{ki}(\balpha^*, \bbeta^*) = c_i$. Substituting these analytical weights into the closed form gives:
% $$\bmz_k(x; \balpha^*, \bbeta^*) = \sum_{i=1}^k c_i\, \bme_i(x) = \bmu.$$
% Because $\|\bmu\|_2 = 1$, we obtain $\|\bmz_k\|_2 = 1 > 0$, ensuring $\hatz_k(x; \balpha^*, \bbeta^*) = \bmu$.
% For the standard Register scheme, no IIR coefficients exist, and therefore, the vector $\hate_k(x) = \bme_k(x)/\|\bme_k(x)\|_2$ is deterministically locked by the tuple $(\btheta, \bmR, x)$.
% \end{proof}
\begin{proof}
Fix $(\btheta, \bmR)$, $k \in \{1, \dots, K\}$, and an input $x$ with $\bV_k(x) \neq \{\bm{0}\}$. Unrolling the recurrence in (\ref{eq:app_closed}) gives
\begin{equation}
\bmz_k(x; \balpha, \bbeta) = \sum_{i=1}^k M_{ki}(\balpha, \bbeta)\, \bme_i(x),
\label{eq:app_closed}
\end{equation}
where $M_{ki}(\balpha, \bbeta) \triangleq \alpha_i \prod_{j=i+1}^k \beta_j$ and the empty product equals $1$, so that $M_{kk} = \alpha_k$.

($\subseteq$).
By (\ref{eq:app_closed}), $\bmz_k(x; \balpha, \bbeta) \in \bV_k(x)$ for all $(\balpha, \bbeta) \in \R^K \times \R^K$. If $\|\bmz_k\|_2 > 0$, then $\hatz_k = \bmz_k / \|\bmz_k\|_2 \in \bV_k(x) \cap \Sph^{D-1}$. Hence $\mathcal{D}^\dep_k(x; \btheta, \bmR) \subseteq \bV_k(x) \cap \Sph^{D-1}$ and $\{\bmz_k(x; \balpha, \bbeta)\} \subseteq \bV_k(x)$.

($\supseteq$).
Let $\bmu \in \bV_k(x)$. By the definition of $\bV_k(x)$, there exist $c_1, \dots, c_k \in \R$ (not necessarily unique) such that $\bmu = \sum_{i=1}^k c_i\, \bme_i(x)$. Choose the coefficients
\begin{equation*}
\begin{aligned}
\balpha^* &\triangleq [c_1, c_2, \dots, c_k, 0, \dots, 0] \in \R^K,\\
\bbeta^* &\triangleq [0, 1, 1, \dots, 1] \in \R^K.
\end{aligned}
\end{equation*}
For $i \le k$, every factor of $\prod_{j=i+1}^k \beta^*_j$ has index $j \ge 2$ and equals $1$, so $M_{ki}(\balpha^*, \bbeta^*) = c_i$; the value of $\beta^*_1$ is irrelevant because $\bmz_0 = \bm{0}$. Substituting into~\eqref{eq:app_closed} gives
$$\bmz_k(x; \balpha^*, \bbeta^*) = \sum_{i=1}^k c_i\, \bme_i(x) = \bmu.$$
Since $\bmu \in \bV_k(x)$ is arbitrary, $\{\bmz_k(x; \balpha, \bbeta)\} \supseteq \bV_k(x)$, which proves the unnormalized claim $\{\bmz_k(x; \balpha, \bbeta)\} = \bV_k(x)$. If, in addition, $\bmu \in \bV_k(x) \cap \Sph^{D-1}$, a set that is nonempty because $\bV_k(x) \neq \{\bm{0}\}$, then $\|\bmz_k\|_2 = \|\bmu\|_2 = 1 > 0$ and $\hatz_k(x; \balpha^*, \bbeta^*) = \bmu$. Hence $\bV_k(x) \cap \Sph^{D-1} \subseteq \mathcal{D}^\dep_k(x; \btheta, \bmR)$.

The unlinked learnable token baseline has no coefficients $(\balpha, \bbeta)$. For fixed $(\btheta, \bmR)$ and $x$, its depth-$k$ output is the single vector $\bme_k(x)$, whose direction is the single point $\hate_k(x) = \bme_k(x) / \|\bme_k(x)\|_2$ whenever $\bme_k(x) \neq \bm{0}$.
\end{proof}

\section{Extended Theoretical Analysis and Deferred Proofs for Section \ref{section_opt_adv}}
\label{app:sec5_3_proofs}

In this section, we provide the detailed analytical derivations, auxiliary lemmas, and the formal proof for the optimization advantage established in Section \ref{section_opt_adv}. 

As there, we fix a baseline checkpoint $(\btheta, \bmR)$ with lift $\beeta_0 \triangleq \beeta_0(\btheta, \bmR)$, and $L_{\bcP}$ denotes the InfoNCE risk. We write $L^\dep_{\bcP}(\beeta) \triangleq L_{\bcP}(\bPhi^\dep_{\beeta})$ and $L^\reg_{\bcP}(\btheta, \bmR) \triangleq L_{\bcP}(\bPhi^\reg_{(\btheta, \bmR)})$, and use the gradients $\bmg^\reg$, $\bmg^\dep$, and $\bmg^\dep_{\bbeta}$ of Sec.~\ref{section_opt_adv}. For a batch $\{(x_i, y_i)\}_{i=1}^N$, $L_R(\bPhi)$ denotes the InfoNCE loss $L_R$ of Section ~\ref{sec:objectives} computed with the head $\bPhi$, so that $L_{\bcP}(\bPhi) = \E_{\bcP^N}[L_R(\bPhi)]$, and $\score^\dep$ denotes the score $\score$ computed with the DepThink head.

% Before proceeding, we formally recall the definition of the orthogonal projector used throughout our analysis:
% \begin{equation}
% \Pperp{\bmu}\, \bmv \;\triangleq\; \bmv - \langle \bmu, \bmv\rangle\, \bmu,
% \qquad \bmu \in \Sph^{D-1},\ \bmv \in \R^D.
% \label{eq:app_proj}
% \end{equation}
% This operator simply removes the component of $\bmv$ that is parallel to $\bmu$, returning the orthogonal component.

% ==========================================
\subsection{Partial Derivatives at the Lift Point}
\label{app:subsec_lift_grads}

We first compute the exact partial derivatives of the DepThink states $\bmz_k$ with respect to the coefficients $(\balpha, \bbeta)$ at $\beeta_0$.

\begin{lemma}[Partial derivatives at the lift]
\label{lem:lift-grads}
Fix an input $x$ and for any $k \in \{2, \dots, K\}$ and $i \in \{1, \dots, K\}$, the partial derivatives of the DepThink state $\bmz_k$ at $\beeta_0$ are
\begin{equation}
\begin{aligned}
\frac{\partial \bmz_k}{\partial \alpha_i}\bigg|_{\beeta_0}
&=
\begin{cases} \bme_k(x), & i = k, \\ \mathbf{0}, & i \neq k, \end{cases}\\[1ex]
\frac{\partial \bmz_k}{\partial \beta_i}\bigg|_{\beeta_0}
&=
\begin{cases} \bme_{k-1}(x), & i = k, \\ \mathbf{0}, & i \neq k. \end{cases}
\end{aligned}
\label{eq:partial-z}
\end{equation}
If, in addition, $\bme_k(x) \neq \bm{0}$, then the $\ell_2$-normalized state $\hatz_k$ satisfies
\begin{equation}
\begin{aligned}
\frac{\partial \hatz_k}{\partial \alpha_k}\bigg|_{\beeta_0} &= \mathbf{0},\\
\frac{\partial \hatz_k}{\partial \beta_k}\bigg|_{\beeta_0}
&= \frac{\Pperp{\hate_k(x)}\,\bme_{k-1}(x)}{\|\bme_k(x)\|_2},
\end{aligned}
\label{eq:partial-zhat}
\end{equation}
and $\partial \hatz_k / \partial \alpha_i|_{\beeta_0} = \partial \hatz_k / \partial \beta_i|_{\beeta_0} = \mathbf{0}$ for $i \neq k$. Here $\Pperp{\hate_k(x)}$ is the orthogonal projection onto the orthogonal complement of $\hate_k(x)$, that is,
\begin{equation}
\begin{aligned}
& \Pperp{\hate_k(x)}\,\bme_{k-1}(x) \\
&= \bme_{k-1}(x) \\
&\quad- \langle \hate_k(x),\bme_{k-1}(x)\rangle\,\hate_k(x) \in \R^D.
\end{aligned}
\label{eq:app_proj}
\end{equation}
For $k = 1$, $\partial \bmz_1 / \partial \alpha_1|_{\beeta_0} = \bme_1(x)$ and $\partial \bmz_1 / \partial \beta_i = \mathbf{0}$ for all $i$, since $\bmz_1 = \beta_1 \bmz_0 + \alpha_1 \bme_1$ with $\bmz_0 = \bm{0}$. If $\bme_1(x) \neq \bm{0}$, all partial derivatives of $\hatz_1$ with respect to $(\balpha, \bbeta)$ vanish at $\beeta_0$.
\end{lemma}

\begin{proof}
By~(\ref{eq:app_closed}), $\bmz_k = \sum_{i=1}^k M_{ki}(\balpha, \bbeta)\, \bme_i$ with $M_{ki} = \alpha_i \prod_{j=i+1}^k \beta_j$. Differentiating gives, for $i, i' \le k$,
\begin{equation}
\begin{aligned}
\frac{\partial \bmz_k}{\partial \alpha_i}
&= \Bigl(\prod_{j=i+1}^k \beta_j\Bigr)\bme_i,\\
\frac{\partial \bmz_k}{\partial \beta_{i'}}
&= \sum_{i<i'} \alpha_i\bme_i
   \prod_{\substack{j=i+1\\j\neq i'}}^k \beta_j,
\end{aligned}
\label{eq:dz-general}
\end{equation}
and both derivatives vanish for $i, i' > k$, since $\bmz_k$ does not depend on $\alpha_i$ or $\beta_{i'}$ for indices beyond $k$. At $\beeta_0$, $\bbeta = \bm{0}$, so a product $\prod_{j \in \mathcal{J}} \beta_j$ is nonzero if and only if the index set $\mathcal{J}$ is empty. For the $\alpha$-derivative, $\{i+1, \dots, k\}$ is empty if and only if $i = k$, which gives $\partial \bmz_k/\partial \alpha_k|_{\beeta_0} = \bme_k(x)$ and zero otherwise. For the $\beta$-derivative, $\{i+1, \dots, k\} \setminus \{i'\}$ with $i < i'$ is empty if and only if $i = k - 1$ and $i' = k$. Since $\alpha_{k-1} = 1$ at $\beeta_0$, this gives $\partial \bmz_k/\partial \beta_k|_{\beeta_0} = \bme_{k-1}(x)$ and zero otherwise, which establishes~(\ref{eq:partial-z}).

Next, the $\ell_2$-normalization $\bmphi(\bmu) \triangleq \bmu/\|\bmu\|_2$ has Jacobian $D\bmphi(\bmu) = \frac{1}{\|\bmu\|_2} \Pperp{\bmu/\|\bmu\|_2}$ at any $\bmu \neq \bm{0}$. Since $\bmz_k|_{\beeta_0} = \bme_k \neq \bm{0}$ and $\bmz_k/\|\bmz_k\|_2 |_{\beeta_0} = \hate_k$, the chain rule gives, for any $\xi \in \{\alpha_i, \beta_i\}_{i=1}^K$,
% \begin{equation}
\begin{align} \notag
\frac{\partial \hatz_k}{\partial \xi}\bigg|_{\beeta_0}
&= D\bmphi(\bmz_k)\big|_{\beeta_0}
   \left[\frac{\partial \bmz_k}{\partial \xi}\bigg|_{\beeta_0}\right]\\
&= \frac{1}{\|\bme_k\|_2}\Pperp{\hate_k}
   \left[\frac{\partial \bmz_k}{\partial \xi}\bigg|_{\beeta_0}\right].
\end{align}
% \end{equation}
For $\xi = \alpha_k$, we substitute $\partial \bmz_k/\partial \alpha_k|_{\beeta_0} = \bme_k(x)$. Since $\bme_k$ is parallel to $\hate_k$, $\Pperp{\hate_k} \bme_k = \mathbf{0}$. For $\xi = \beta_k$, we substitute $\partial \bmz_k/\partial \beta_k|_{\beeta_0} = \bme_{k-1}(x)$, which gives~(\ref{eq:partial-zhat}). For all other $\xi$, the inner derivative is zero by~(\ref{eq:partial-z}). The case $k = 1$ follows in the same way, because $\partial \bmz_1/\partial \beta_i = \mathbf{0}$ and $\Pperp{\hate_1} \bme_1 = \mathbf{0}$.
\end{proof}

Using Lemma~\ref{lem:lift-grads}, we can now compute the closed-form expression for the gradient of the empirical InfoNCE risk with respect to $\beta_k$.

\begin{theorem}[Closed-form empirical InfoNCE gradient at the lift]
\label{thm:beta-grad}
Let $\{(x_i, y_i)\}_{i=1}^N$ be a batch such that $\bme_k(x_i) \neq \bm{0}$ and $\bme_k(y_j) \neq \bm{0}$ for all $i, j \in \{1, \dots, N\}$ and $k \in \{1, \dots, K\}$. Let $\score^\dep(x, y) \triangleq \frac{1}{K}\sum_{k'=1}^K \langle \hatz_{k'}(x), \hatz_{k'}(y)\rangle$ denote the score $\score$ defined in Section ~\ref{sec:objectives} which is computed with the DepThink head, and let $p_{ij} \triangleq \exp(\score^\dep(x_i, y_j)/\tau)/\sum_{l=1}^N \exp(\score^\dep(x_i, y_l)/\tau)$ denote the InfoNCE softmax probabilities of the DepThink score. By Proposition~\ref{prop:lift-id}, they coincide at $\beeta_0$ with the probabilities $p_{ij}^\reg$ of the unlinked learnable token baseline (computed with $\hate_{k'}$ in place of $\hatz_{k'}$), that is, $p_{ij}|_{\beeta_0} = p_{ij}^\reg$. For any input $x$, let
\begin{equation*}
\bm{\rho}_k(x) \triangleq \frac{\Pperp{\hate_k(x)}\,\bme_{k-1}(x)}{\|\bme_k(x)\|_2}
\end{equation*}
denote the part of $\bme_{k-1}(x)$ orthogonal to $\hate_k(x)$, scaled by $1/\|\bme_k(x)\|_2$, which equals $\partial \hatz_k(x)/\partial \beta_k|_{\beeta_0}$ by Lemma~\ref{lem:lift-grads}. For any $k \in \{2, \dots, K\}$,
\begin{align}
\tau \frac{\partial L_R(\bPhi^\dep)}{\partial \beta_k}\bigg|_{\beeta_0}
&= \frac{1}{NK}\sum_{i,j=1}^N(p_{ij}-\delta_{ij}) \notag\\
&\quad \cdot \bigl[\langle \bm{\rho}_k(x_i), \hate_k(y_j)\rangle \notag\\
&\qquad + \langle \hate_k(x_i), \bm{\rho}_k(y_j)\rangle\bigr],
\label{eq:beta-grad}
\end{align}
where $\delta_{ij}$ is the Kronecker delta, and $\partial L_R(\bPhi^\dep)/\partial \beta_1|_{\beeta_0} = 0$.
\end{theorem}
\begin{proof}
For a given query index $i$, the per-query InfoNCE loss is
\begin{equation*}
\begin{aligned} 
q_i & \triangleq -\log p_{ii} \\
&= -\frac{\score^\dep(x_i,y_i)}{\tau}\\
&\quad+\log\sum_{l=1}^N
\exp\bigl(\score^\dep(x_i,y_l)/\tau\bigr).
\end{aligned}
\end{equation*}
By the standard softmax cross-entropy identity, the partial derivative of $q_i$ with respect to a single score $\score^\dep(x_i, y_j)$ is
\begin{equation}
\begin{aligned}
\frac{\partial q_i}{\partial \score^\dep(x_i,y_j)}
&= -\frac{\delta_{ij}}{\tau}
 +\frac{1}{\tau}
 \frac{\exp(\score^\dep(x_i,y_j)/\tau)}
 {\sum_l\exp(\score^\dep(x_i,y_l)/\tau)}\\
&= \frac{1}{\tau}(p_{ij}-\delta_{ij}).
\end{aligned}
\label{eq:cee-grad}
\end{equation}
Since $q_{i'}$ depends only on the scores $\{\score^\dep(x_{i'}, y_l)\}_{l = 1}^N$, $\partial q_{i'}/\partial \score^\dep(x_i, y_j) = 0$ for all $i' \neq i$. As $ L_R(\bPhi^\dep) = N^{-1} \sum_{i'} q_{i'}$, we obtain
% \begin{equation}
\begin{align} \notag
\frac{\partial  L_R(\bPhi^\dep)}{\partial \score^\dep(x_i,y_j)}
&= \frac{1}{N}\sum_{i'=1}^N
\frac{\partial q_{i'}}{\partial \score^\dep(x_i,y_j)}\\ \notag
&= \frac{1}{N}\frac{\partial q_i}{\partial \score^\dep(x_i,y_j)} \\ \label{eq:loss-score}
& = \frac{p_{ij}-\delta_{ij}}{N\tau}.
\end{align}
% \end{equation}
Differentiating the score $\score^\dep(x, y) = \frac{1}{K}\sum_{k'=1}^K \langle \hatz_{k'}(x), \hatz_{k'}(y)\rangle$ by the product rule gives
\begin{align} \notag
& \frac{\partial \score^\dep(x_i,y_j)}{\partial \beta_k} \\ \notag
& = \frac{1}{K}\sum_{k'=1}^K \Biggl(
\Bigl\langle \frac{\partial \hatz_{k'}(x_i)}{\partial \beta_k},
\hatz_{k'}(y_j)\Bigr\rangle\\ \label{eq:score-chain-full}
& \quad +\Bigl\langle \hatz_{k'}(x_i),
\frac{\partial \hatz_{k'}(y_j)}{\partial \beta_k}\Bigr\rangle\Biggr).
\end{align}
We evaluate this expression at $\beeta_0$, where $\hatz_{k'} = \hate_{k'}$. By Lemma~\ref{lem:lift-grads}, the derivatives $\partial \hatz_{k'}(x)/\partial \beta_k$ are sparse, with
\begin{equation}
\frac{\partial \hatz_{k'}(x)}{\partial \beta_k}\bigg|_{\beeta_0} \;=\;
\begin{cases}
\bm{\rho}_k(x), & k' = k, \\
\mathbf{0}, & k' \neq k,
\end{cases}
\label{eq:zhat-beta-cases}
\end{equation}
for every query or target $x$. Substituting into~(\ref{eq:score-chain-full}), only the term $k' = k$ remains, which gives
\begin{align} \notag
\frac{\partial \score^\dep(x_i,y_j)}{\partial \beta_k}\bigg|_{\beeta_0}
&= \frac{1}{K}\bigl[\langle \bm{\rho}_k(x_i), \hate_k(y_j)\rangle\\ \label{eq:score-beta}
&\qquad + \langle \hate_k(x_i), \bm{\rho}_k(y_j)\rangle\bigr].
\end{align}
Finally, the chain rule with~(\ref{eq:loss-score}) and~(\ref{eq:score-beta}), multiplied by $\tau$, gives
\begin{align}
& \tau \frac{\partial L_R(\bPhi^\dep)}{\partial \beta_k}\bigg|_{\beeta_0} \\
&= \tau\sum_{i,j=1}^N
\frac{\partial L_R(\bPhi^\dep)}{\partial \score^\dep(x_i,y_j)}  \cdot \frac{\partial \score^\dep(x_i,y_j)}{\partial \beta_k}\bigg|_{\beeta_0}\notag\\
&= \frac{1}{NK}\sum_{i,j=1}^N(p_{ij}-\delta_{ij}) \notag\\
&\qquad \cdot \bigl[\langle \bm{\rho}_k(x_i), \hate_k(y_j)\rangle \notag\\
&\quad \quad + \langle \hate_k(x_i), \bm{\rho}_k(y_j)\rangle\bigr].
\end{align}
For $k = 1$, all derivatives of $\hatz_{k'}$ with respect to $\beta_1$ vanish at $\beeta_0$ by Lemma~\ref{lem:lift-grads}, so $\partial L_R(\bPhi^\dep)/\partial \beta_1|_{\beeta_0} = 0$.
\end{proof}

% ==========================================
\subsection{Gradient Identities and Descent Gap Expansion}
\label{app:subsec_gap_expansion}

To analyze the gradient descent dynamics, we first make a standard local smoothness assumption required for bounding the descent trajectory.

\begin{assumption}[Local regularity near the lift]
\label{ass:smooth_app}
Let $U_0$ be the closed ball of radius $r_{U_0} > 0$ around $\beeta_0$, that is, the set of all $\beeta$ with $\|\beeta - \beeta_0\|_2 \le r_{U_0}$.
\begin{enumerate}[label=(\alph*),nosep,leftmargin=2em]
\item The token hidden states are almost surely nonzero, that is, $\bme_k(x) \neq \bm{0}$ and $\bme_k(y) \neq \bm{0}$ for all $k \in \{1, \dots, K\}$ and $\bcP$-almost every pair $(x, y)$.
\item $L^\dep_{\bcP}$ is differentiable on $U_0$, and its gradient is $L_{\mathrm{loc}}$-Lipschitz there, so that
\begin{align} \notag
& \|\nabla L^\dep_{\bcP}(\beeta)-\nabla L^\dep_{\bcP}(\beeta')\|_2 \\
&\le L_{\mathrm{loc}}\|\beeta-\beeta'\|_2  \quad  \text{for all }\beeta,\beeta'\in U_0.
\end{align}
\item The gradient at $\beeta_0$ can be taken under the expectation, so that $\nabla L^\dep_{\bcP}(\beeta_0) = \E_{\bcP^N}\bigl[\nabla_{\beeta} L_R(\bPhi^\dep_{\beeta})|_{\beeta_0}\bigr]$.
\end{enumerate}
\end{assumption}
Assumption~\ref{ass:smooth_app} is local, since it constrains the risk only on a ball around the checkpoint and does not require global smoothness. Part (a) makes the normalization differentiable at $\beeta_0$, where $\bmz_k = \bme_k$, so Lemma~\ref{lem:lift-grads} and Theorem~\ref{thm:beta-grad} apply almost surely. Together with part (c), Theorem~\ref{thm:beta-grad} then gives $\bmg^\dep_{\bbeta} = \E_{\bcP^N}[\nablab L_R(\bPhi^\dep)|_{\beeta_0}]$, whose $k$-th entry is the expectation of~(\ref{eq:beta-grad}). Moreover, Proposition~\ref{prop:lift-id} gives $L^\reg_{\bcP}(\btheta', \bmR') = L^\dep_{\bcP}(\btheta', \bmR', \bm{1}, \bm{0})$ for all $(\btheta', \bmR')$. Hence $\nabla L^\reg_{\bcP}(\btheta', \bmR')$ is the $(\btheta, \bmR)$-block of $\nabla L^\dep_{\bcP}(\btheta', \bmR', \bm{1}, \bm{0})$, and part (b) implies that $\nabla L^\reg_{\bcP}$ is also $L_{\mathrm{loc}}$-Lipschitz on the ball of radius $r_{U_0}$ around $(\btheta, \bmR)$.

% \begin{assumption}[Local smoothness near the lift]
% \label{ass:smooth_app}
% For any given backbone parameters $(\btheta, \bmR) \in \bTheta \times \R^{K \times D}$, there exists an open neighborhood $U_0$ around the lift point $\beeta_0 \triangleq (\btheta, \bmR, \bm{1}, \bm{0})$ where the expected InfoNCE risk $L_{\bcP}$ is twice continuously differentiable and its gradient is Lipschitz continuous with constant $L_{\mathrm{loc}} > 0$, that is,
% \begin{equation}
% \begin{aligned}
% \|\nabla L_{\bcP}(\beeta)-\nabla L_{\bcP}(\beeta')\|_2
% &\le L_{\mathrm{loc}}\|\beeta-\beeta'\|_2,\\
% &\hspace{-4em}\text{for all }\beeta,\beeta'\in U_0.
% \end{aligned}
% \end{equation}
% \end{assumption}
% Assumption \ref{ass:smooth_app} is a standard local non-convex regularity assumption which is widely adopted
% in the theoretical optimization literature. 
% % It is worth mentioning that we only require local Lipschitz smoothness rather than  global Lipschitz smoothness.
% % which is known to fail for transformer losses
% % (\cite{zhang2020clipping, cohen2021edgeOfStability}).

Before comparing the gradient steps, we first provide a lemma that reveals the relationships between the gradient of DepThink and that of unlinked learnable token baseline at the lift.

% \begin{lemma}[Lift Gradient Identity]
% \label{lem:lift-grad-id}
% At the lift point $\beeta_0(\btheta, \bmR)$, the following identities hold
% \begin{enumerate}[label=(\roman*),nosep,leftmargin=2em]
% \item $L_{\bcP}(\bPhi^\dep_{\beeta_0}) = L_{\bcP}(\bPhi^\reg_{(\btheta, \bmR)})$.
% \item $\nabla_{(\btheta, \bmR)} L_{\bcP}(\bPhi^\dep)|_{\beeta_0} = \nabla_{(\btheta, \bmR)} L_{\bcP}(\bPhi^\reg)|_{(\btheta, \bmR)}$.
% \item $\nabla_{\balpha} L_{\bcP}(\bPhi^\dep)|_{\beeta_0} = \mathbf{0}$.
% \item $\|\nabla L_{\bcP}(\bPhi^\dep)|_{\beeta_0}\|^2 = \|\nabla_{(\btheta, \bmR)} L_{\bcP}(\bPhi^\reg)|_{(\btheta, \bmR)}\|^2 + \|\nablab L_{\bcP}(\bPhi^\dep)|_{\beeta_0}\|^2$.
% \end{enumerate}
% \end{lemma}

\begin{lemma}[Lift Gradient Identity]
\label{lem:lift-grad-id}
Under Assumption~\ref{ass:smooth_app}(a), the following identities hold at the lift $\beeta_0(\btheta, \bmR)$.
\begin{enumerate}[label=(\roman*),nosep,leftmargin=2em]
\item $L_{\bcP}(\bPhi^\dep_{\beeta_0}) = L_{\bcP}(\bPhi^\reg_{(\btheta, \bmR)})$.
\item $\nabla_{(\btheta, \bmR)} L_{\bcP}(\bPhi^\dep)|_{\beeta_0}$\newline
$= \nabla_{(\btheta, \bmR)} L_{\bcP}(\bPhi^\reg)|_{(\btheta, \bmR)}$.
\item $\nabla_{\balpha} L_{\bcP}(\bPhi^\dep)|_{\beeta_0} = \bm{0}$.
\item $\|\nabla L_{\bcP}(\bPhi^\dep)|_{\beeta_0}\|^2$\newline
$= \|\nabla_{(\btheta, \bmR)} L_{\bcP}(\bPhi^\reg)|_{(\btheta, \bmR)}\|^2$\newline
$\qquad \qquad + \|\nablab L_{\bcP}(\bPhi^\dep)|_{\beeta_0}\|^2$.
\end{enumerate}
\end{lemma}

\begin{proof}
Identity (i) is a direct restatement of Proposition 1. 

For (ii), note that restricting the function $L_{\bcP}^\dep(\btheta, \bmR, \balpha, \bbeta)$ to the condition where $\balpha = \bm{1}$ and $\bbeta = \bm{0}$ perfectly recovers $L_{\bcP}^\reg(\btheta, \bmR)$. Since the partial derivatives with respect to $(\btheta, \bmR)$ are taken strictly along this condition, they are equal.

To prove (iii), we consider a neighborhood where $\balpha > \mathbf{0}$ and $\bbeta = \bm{0}$. In this region, the latent thought $\bmz_k$  simplifies to $\bmz_k = \alpha_k \bme_k$. Because the $\ell_2$-normalization is scale-invariant, we have $\hatz_k = (\alpha_k \bme_k) / \|\alpha_k \bme_k\|_2 = \hate_k$ for any $\alpha_k > 0$. Thus, the normalized $\hatz_k$ and the expected risk $L_{\bcP}^\dep$ are completely independent of $\balpha$ near the lift point. Therefore, the gradient with respect to $\balpha$ is zero, that is, $\nabla_{\balpha} L_{\bcP}(\bPhi^\dep)|_{\beeta_0} = \mathbf{0}$.

Finally, (iv) holds because the parameter groups $(\btheta, \bmR, \balpha, \bbeta)$ are independent, meaning the squared $\ell_2$-norm of the full gradient decomposes into the sum of the squared norms of its parts, that is, $\|\nabla L_{\bcP}^\dep\|^2 = \|\nabla_{(\btheta, \bmR)} L_{\bcP}^\dep\|^2 + \|\nabla_{\balpha} L_{\bcP}^\dep\|^2 + \|\nablab L_{\bcP}^\dep\|^2$. Substituting (ii) and (iii) directly yields the result.
\end{proof}

We now  develop the closed-form expression for the descent gap after a single gradient step.

\begin{lemma}[Descent Gap Expansion; Lemma~\ref{lem:gap-expansion_main} restated]
\label{lem:gap-expansion}
Suppose Assumption~\ref{ass:smooth_app} holds. Let $(\btheta', \bmR') \triangleq (\btheta, \bmR) - \mu \bmg^\reg$ and $\beeta' \triangleq \beeta_0 - \mu \bmg^\dep$ be the parameters after one gradient step of size $\mu > 0$, and define
\begin{equation}
\mu_{\max} \triangleq \min \left( \frac{1}{L_{\mathrm{loc}}}, \, \frac{r_{U_0}}{\|\bmg^\dep\|}, \, \frac{r_{U_0}}{\|\bmg^\reg\|} \right),
\label{eq:mu-max}
\end{equation}
with $r_{U_0}/0 \triangleq \infty$, which keeps both updates in $U_0$. For the one-step descent gap between the unlinked learnable token baseline and DepThink,
\begin{equation}
\Delta(\mu) \;\triangleq\; L_{\bcP}\bigl(\bPhi^\reg_{(\btheta', \bmR')}\bigr) - L_{\bcP}\bigl(\bPhi^\dep_{\beeta'}\bigr).
\label{eq:gap-def-appendix}
\end{equation}
For any $0 < \mu \le \mu_{\max}$, we have
\begin{equation}
\Delta(\mu)  = \mu \|\bmg^\dep_{\bbeta}\|^2 + O(\mu^2),
\label{eq:gap-expansion}
\end{equation}
with $|O(\mu^2)| \le \frac{\mu^2}{2} L_{\mathrm{loc}} \bigl(\|\bmg^\reg\|^2 + \|\bmg^\dep\|^2\bigr)$.
\end{lemma}

\begin{proof}
For DepThink, the fundamental theorem of calculus along the segment from $\beeta_0$ to $\beeta'$ gives
\begin{align} \notag
& L^\dep_{\bcP}(\beeta_0 - \mu \bmg^\dep) \\ \label{dep_grad_expansion}
& = L^\dep_{\bcP}(\beeta_0) - \mu \|\bmg^\dep\|^2 + \varepsilon^\dep,
\end{align}
where the remainder is
\begin{equation}
\begin{aligned}
\varepsilon^\dep \triangleq \int_0^\mu \bigl\langle & \nabla L^\dep_{\bcP}(\beeta_0 - t \bmg^\dep) \\
& - \nabla L^\dep_{\bcP}(\beeta_0),\, -\bmg^\dep \bigr\rangle \, dt.
\end{aligned}
\end{equation}
Since $\mu \le \mu_{\max}$, every point $\beeta_0 - t \bmg^\dep$ with $t \in [0, \mu]$ satisfies $\|t \bmg^\dep\| \le r_{U_0}$ and lies in $U_0$. The Cauchy-Schwarz inequality and Assumption~\ref{ass:smooth_app}(b) then give
\begin{align} \notag
|\varepsilon^\dep| &\le \int_0^\mu \|\bmg^\dep\|_2 \, \|\nabla L^\dep_{\bcP}(\beeta_0 - t \bmg^\dep) \\ \notag
&\qquad\qquad - \nabla L^\dep_{\bcP}(\beeta_0)\|_2 \, dt \\ \notag
&\le \int_0^\mu L_{\mathrm{loc}} \, t \, \|\bmg^\dep\|_2^2 \, dt  \\ \label{dep_residual}
& = \frac{\mu^2}{2} L_{\mathrm{loc}} \|\bmg^\dep\|^2.
\end{align}
For the unlinked learnable token baseline, the points $(\btheta, \bmR) - t\bmg^\reg$ with $t \in [0, \mu]$ lie within distance $r_{U_0}$ of $(\btheta, \bmR)$, where $\nabla L^\reg_{\bcP}$ is $L_{\mathrm{loc}}$-Lipschitz (see the discussion after Assumption~\ref{ass:smooth_app}). The same argument gives
\begin{align} \notag
    & rL^\reg_{\bcP}((\btheta, \bmR) - \mu \bmg^\reg) \\  \label{reg_grad_expansion} 
    & = L^\reg_{\bcP}(\btheta, \bmR) - \mu \|\bmg^\reg\|^2 + \varepsilon^\reg,
\end{align}
with
\begin{equation} \label{reg_residual}
    |\varepsilon^\reg| \le \frac{\mu^2}{2} L_{\mathrm{loc}} \|\bmg^\reg\|^2.
\end{equation}
Subtracting~(\ref{dep_grad_expansion}) from~(\ref{reg_grad_expansion}) and using $L^\dep_{\bcP}(\beeta_0) = L^\reg_{\bcP}(\btheta, \bmR)$ from Lemma~\ref{lem:lift-grad-id}(i), we obtain
\begin{align*}
\Delta(\mu) &= \mu\, \bigl(\|\bmg^\dep\|^2 - \|\bmg^\reg\|^2\bigr) + (\varepsilon^\reg - \varepsilon^\dep) \\
&= \mu\, \|\bmg^\dep_{\bbeta}\|^2 + (\varepsilon^\reg - \varepsilon^\dep),
\end{align*}
where the second line uses Lemma~\ref{lem:lift-grad-id}(iv). The remainder bound follows from the triangle inequality,
\begin{equation}
\begin{aligned}
|\varepsilon^\reg - \varepsilon^\dep| &\le |\varepsilon^\reg| + |\varepsilon^\dep| \\
&\le \frac{\mu^2}{2} L_{\mathrm{loc}} \bigl(\|\bmg^\reg\|^2 + \|\bmg^\dep\|^2\bigr),
\end{aligned}
\end{equation}
which completes the proof.
% The factor $1/L_{\mathrm{loc}}$ in $\mu_{\max}$ is not needed for this bound. It keeps both steps in the standard descent regime, since for $\mu \le 1/L_{\mathrm{loc}}$ the same expansion gives $L^\dep_{\bcP}(\beeta') \le L^\dep_{\bcP}(\beeta_0) - \frac{\mu}{2}\|\bmg^\dep\|^2$, and likewise for the baseline.
\end{proof}

% ==========================================
\subsection{Proof of Theorem \texorpdfstring{{\ref{thm:1step}}}{\ref{thm:1step}}: Weak and Strict Optimization Advantage}
\label{app:subsec_proof_theorem1}

We are now ready to prove the main theorem presented in Section {\ref{section_opt_adv}}, restated here for convenience.

\textbf{Restatement of Theorem~\ref{thm:1step}.}
Under the assumptions of Lemma~\ref{lem:gap-expansion_main}, define
\begin{equation}
\bar{\mu} \;\triangleq\; \frac{2\|\bmg^\dep_{\bbeta}\|^2}{L_{\mathrm{loc}} \bigl(\|\bmg^\reg\|^2 + \|\bmg^\dep\|^2\bigr)}.
\label{eq:mu-bar-appendix}
\end{equation}
Then, for every step size covered by the two cases below, $\Delta(\mu) \ge 0$ (the DepThink step is never worse than the baseline step), with equality if and only if $\bmg^\dep_{\bbeta} = \bm{0}$.
\begin{enumerate}[label=(\roman*), leftmargin=2em, nosep]
\item \textbf{Weak advantage.} If $\bmg^\dep_{\bbeta} = \bm{0}$, then $\beeta' = \beeta_0(\btheta', \bmR')$ and $\Delta(\mu) = 0$ for every $\mu > 0$.
\item \textbf{Strict advantage.} If $\bmg^\dep_{\bbeta} \neq \bm{0}$, then $\Delta(\mu) > 0$ for every $\mu \in (0, \min(\bar{\mu}, \mu_{\max}))$.
\end{enumerate}

% \textbf{Restatement of Theorem \ref{thm:1step}}
% Suppose the expected InfoNCE risk satisfies the local smoothness Assumption \ref{ass:smooth_app} in Appendix~\ref{app:sec5_3_proofs} with Lipschitz constant $L_{\mathrm{loc}} > 0$, admitting a maximum valid step size $\mu_{\max} > 0$ that confines the update to the locally Lipschitz neighborhood. Define the critical step-size threshold
% \begin{equation}
% \bar{\mu} \;\triangleq\; \frac{2\bigl\|\nablab L_{\bcP}(\bPhi^\dep)|_{\beeta_0}\bigr\|^2}{L_{\mathrm{loc}}\, \bigl(\|\bmg^\reg\|^2 + \|\bmg^\dep\|^2\bigr)}.
% \label{eq:mu-bar-appendix}
% \end{equation}
% Then, for any gradient descent step size $\mu \in \bigl(0, \min(\bar{\mu}, \mu_{\max})\bigr)$, the following properties hold
% \begin{enumerate}[leftmargin=2em,nosep]
% \item \textbf{Weak Optimization Advantage:} $\Delta(\mu) \ge 0$ whenever $\mu \in \bigl(0, \min(\bar{\mu}, \mu_{\max})\bigr)$. Furthermore, the exact equality $\Delta(\mu) = 0$ holds {if and only if} $\nablab L_{\bcP}(\bPhi^\dep)|_{\beeta_0} = \mathbf{0}$, in which case the descent trajectories of {DepThink} and {Register} perfectly coincide.
% \item \textbf{Strict Optimization Advantage:} If $\nablab L_{\bcP}(\bPhi^\dep)|_{\beeta_0} \neq \mathbf{0}$, then $\Delta(\mu) > 0$ strictly for all $\mu \in \bigl(0, \min(\bar{\mu}, \mu_{\max})\bigr)$, guaranteeing that {DepThink} consistently achieves a strictly greater InfoNCE risk reduction than {Register}.
% \end{enumerate}

\begin{table*}[t]
\centering
\small
\caption{Model scaling on multimodal retrieval across training paradigms, model sizes, and dense/sparse architectures.
}
\label{app:tab:scale_architecture}
\begin{tabular}{llcccc}
\toprule
Setting & Method & CIRR & MSCOCO T2I & MSVD & QVHighlights \\
% \midrule
% 2B, LoRA & DepThink & 65.5 & 79.3 & 77.2 & 67.9 \\
\midrule
\multirow{3}{*}{2B, full FT}
& LoopLM & 61.7 & 78.1 & 76.6 & 68.1 \\
& \textbf{DepThink} & \textbf{67.9} & \textbf{79.4} & \textbf{78.4} & \textbf{71.1} \\
& \quad (-residual) & 64.2 & 76.6 & 76.0 & 65.9 \\
\midrule
\multirow{3}{*}{8B, LoRA}
& LoopLM & 67.2 & 78.1 & 85.8 & 67.8 \\
& \textbf{DepThink} & \textbf{70.0} & \textbf{80.1} & \textbf{86.7} & \textbf{74.4} \\
& \quad (-residual) & 68.0 & 79.3 & 85.4 & 72.8 \\
\midrule
\multirow{3}{*}{\begin{tabular}[c]{@{}l@{}}30B-A3B MoE,\\LoRA\end{tabular}}
& LoopLM & 70.5 & 80.4 & 83.4 & 67.5 \\
& \textbf{DepThink} & \textbf{73.3} & \textbf{81.3} & \textbf{84.0} & \textbf{71.5} \\
& \quad (-residual) & 72.3 & 79.4 & 82.8 & 69.2 \\
\bottomrule
\end{tabular}
\end{table*}

\begin{proof}
(i) Suppose $\bmg^\dep_{\bbeta} = \bm{0}$. By Lemma~\ref{lem:lift-grad-id}(ii) and (iii), $\bmg^\dep = (\bmg^\reg, \bm{0}, \bm{0})$. For any $\mu > 0$, the DepThink step therefore updates $(\btheta, \bmR)$ exactly as the unlinked learnable token baseline step does and leaves $(\balpha, \bbeta) = (\bm{1}, \bm{0})$ unchanged, so $\beeta' = (\btheta', \bmR', \bm{1}, \bm{0}) = \beeta_0(\btheta', \bmR')$ is the lift of the updated baseline parameters. By Proposition~\ref{prop:lift-id}, $\bPhi^\dep_{\beeta'} = \bPhi^\reg_{(\btheta', \bmR')}$ and the two risks are equal, so $\Delta(\mu) = 0$ for every $\mu > 0$. This case needs no step-size restriction, and the two trajectories coincide whenever the coefficients $\bbeta$ receive no gradient signal.

(ii) Suppose $\bmg^\dep_{\bbeta} \neq \bm{0}$. Then $\|\bmg^\reg\|^2 + \|\bmg^\dep\|^2 \ge \|\bmg^\dep_{\bbeta}\|^2 > 0$ by Lemma~\ref{lem:lift-grad-id}(iv), so $\bar{\mu}$ is well defined and positive. For $\mu \le \mu_{\max}$, the expansion and remainder bound of Lemma~\ref{lem:gap-expansion} give
\begin{align} \notag
\Delta(\mu) \ge & \mu \|\bmg^\dep_{\bbeta}\|^2 \\
& \;\; - \frac{\mu^2}{2} L_{\mathrm{loc}} \bigl(\|\bmg^\reg\|^2 + \|\bmg^\dep\|^2\bigr).
\end{align}
The right-hand side is positive if and only if the linear gain exceeds the quadratic penalty. Dividing by $\mu > 0$, this is equivalent to
\begin{equation}
    \mu < \frac{2\|\bmg^\dep_{\bbeta}\|^2}{L_{\mathrm{loc}}\, \bigl(\|\bmg^\reg\|^2 + \|\bmg^\dep\|^2\bigr)} = \bar{\mu}.
\end{equation}
Combining this with the requirement $\mu \le \mu_{\max}$ of Lemma~\ref{lem:gap-expansion}, $\Delta(\mu) > 0$ for every $\mu \in (0, \min(\bar{\mu}, \mu_{\max}))$.

Cases (i) and (ii) together show that $\Delta(\mu) \ge 0$ on the stated step sizes, with equality exactly in case (i), that is, if and only if $\bmg^\dep_{\bbeta} = \bm{0}$.
\end{proof}

\raggedbottom
\section{Prompts for LLMs as Judges}
\label{sec:prompts-llm-judges}

We use Gemini-2.5-Pro as the judge with the system prompt and scoring rubric shown in Fig.~\ref{fig:judge_rubric}.

\tcbset{
  rubricbox/.style={
    colback=white,
    colframe=blue!70!black,
    fonttitle=\bfseries,
    coltitle=white,
    colbacktitle=blue!70!black,
    boxrule=0.8pt,
    arc=1pt,
    left=8pt, right=8pt, top=6pt, bottom=6pt,
    title filled
  }
}

\begin{figure*}[t]
%% breakable
\begin{tcolorbox}[rubricbox,title={LLM-as-a-Judge Evaluation Rubric}]

\textbf{System Prompt:}\\
\textit{You are an expert evaluator specializing in assessing the quality of Chain-of-Thought (COT) reasoning for visual understanding tasks. You will be given an image, a task description, a ground-truth reference answer, and a COT reasoning trace. Your job is to score the COT on two dimensions: Semantic Accuracy and Coherence.}

\vspace{10pt}
\textbf{1. Semantic Accuracy (1--5)}\\
Does the COT correctly capture the essential meaning, facts, and logical relationships needed to solve the task? Does it arrive at (or support) the correct answer as given by the ground truth?

\vspace{6pt}
\begin{tabular}{@{}c p{0.85\textwidth}@{}}
\toprule
\textbf{Score} & \textbf{Description} \\
\midrule
5 & All key facts, entities, and relationships are correctly identified. The reasoning leads to the correct answer consistent with the ground truth. No factual errors or hallucinations. \\[4pt]
4 & Most key information is correct with only minor omissions. The final answer aligns with the ground truth. No significant factual errors. \\[4pt]
3 & Core meaning is partially captured but some important details are missing or slightly inaccurate. The answer may partially match the ground truth. \\[4pt]
2 & Multiple factual errors or significant omissions. The reasoning leads to an answer that deviates substantially from the ground truth. \\[4pt]
1 & Mostly incorrect, irrelevant, or hallucinated content. The conclusion contradicts the ground truth. \\
\bottomrule
\end{tabular}

\vspace{10pt}
\textbf{2. Coherence (1--5)}\\
Is the COT logically structured, internally consistent, and easy to follow as a reasoning chain?

\vspace{6pt}
\begin{tabular}{@{}c p{0.85\textwidth}@{}}
\toprule
\textbf{Score} & \textbf{Description} \\
\midrule
5 & Clear logical flow from observation to conclusion. Each step follows naturally from the previous one. Well-organized and concise. \\[4pt]
4 & Generally logical with minor gaps or redundancies. The overall reasoning direction is clear. \\[4pt]
3 & Partially coherent but contains noticeable logical jumps, contradictions, or disorganized segments. \\[4pt]
2 & Fragmented reasoning with weak connections between steps. Hard to follow the logical chain. \\[4pt]
1 & Incoherent, self-contradictory, or completely disorganized. No discernible reasoning structure. \\
\bottomrule
\end{tabular}

\end{tcolorbox}
\caption{System prompt and scoring rubric for the LLM-as-a-judge evaluation.}
\label{fig:judge_rubric}
% \vspace{30pt}

\end{figure*}

\section{Additional Empirical Results}
\label{app:additional_results}

\begin{figure}[t]
\centering
\begin{minipage}[t]{0.32\linewidth}
    \vspace{0pt}
    \centering
    \begin{tikzpicture}
    \begin{axis}[
        width=0.78\linewidth,
        height=3.6cm,
        scale only axis,
        % Axis & tick styling
        axis line style={draw=black!80, line width=0.6pt},
        tick style={draw=black!80, line width=0.5pt},
        major tick length=2pt,
        tick label style={font=\scriptsize, color=black!85, inner sep=1pt},
        label style={font=\footnotesize, color=black!90},
        % X-axis configuration (log2 scale: K doubles, so points are evenly spaced)
        xmode=log, log basis x=2,
        xlabel={number of special tokens},
        xlabel style={yshift=3pt},
        xmin=1.6, xmax=20,
        xtick={2, 4, 8, 16},
        xticklabels={2, 4, 8, 16},
        xminorticks=false,
        ylabel={Retrieval recall},
        ylabel style={yshift=-2pt},
        ymin=0.42, ymax=0.70,
        ytick={0.45, 0.50, 0.55, 0.60, 0.65, 0.70},
        yticklabel style={
            /pgf/number format/fixed,
            /pgf/number format/precision=2,
            /pgf/number format/zerofill,
        },
        % Subtle grid
        grid=major,
        grid style={draw=gridgray, line width=0.45pt},
        clip=false,
        % Image CLS = solid + filled squares; video RET = dashed + open squares
        icls/.style={line width=1.0pt, mark=square*, mark size=1.6pt,
                     color=#1, mark options={fill=#1, solid}},
        vret/.style={line width=1.0pt, dashed, mark=square*, mark size=1.6pt,
                     color=#1, mark options={fill=white, solid}},
        % Two-entry legend (methods) in the empty band between the groups
        legend cell align=left,
        legend style={
            font=\scriptsize,
            draw=black!50,
            line width=0.4pt,
            fill=white,
            fill opacity=0.85,
            text opacity=1,
            inner sep=1pt,
            row sep=-1.5pt,
            at={(0.5, 0.5)},
            anchor=center,
        },
        legend image code/.code={
            \draw[mark repeat=2, mark phase=2]
                plot coordinates {(0cm,0cm) (0.16cm,0cm) (0.32cm,0cm)};
        },
    ]
        % --- Image classification (solid) ---
        \addplot[icls=depred] coordinates {(2, 0.600) (4, 0.669) (8, 0.676) (16, 0.681)};
        \addlegendentry{DepThink}
        \addplot[icls=resblue] coordinates {(2, 0.600) (4, 0.667) (8, 0.674) (16, 0.675)};
        \addlegendentry{w/o residual}

        % --- Video retrieval (dashed) ---
        \addplot[vret=depred, forget plot] coordinates {(2, 0.444) (4, 0.471) (8, 0.475) (16, 0.483)};
        \addplot[vret=resblue, forget plot] coordinates {(2, 0.438) (4, 0.455) (8, 0.465) (16, 0.472)};

        % Task labels next to each line group
        \node[font=\scriptsize\itshape, text=black!70] at (axis cs:8, 0.645) {Image CLS};
        \node[font=\scriptsize\itshape, text=black!70] at (axis cs:8, 0.497) {Video RET};
    \end{axis}
    \end{tikzpicture}
    \caption{Video RET. vs Image CLS. The retrieval scores are from MMEB-v2 benchmark.}
    \label{fig:video_easy_hard}
\end{minipage}\hfill
\begin{minipage}[t]{0.65\linewidth}
    \vspace{0pt}
    \centering
    \begin{subfigure}[b]{0.49\linewidth}
        \centering
        \includegraphics[width=\linewidth]{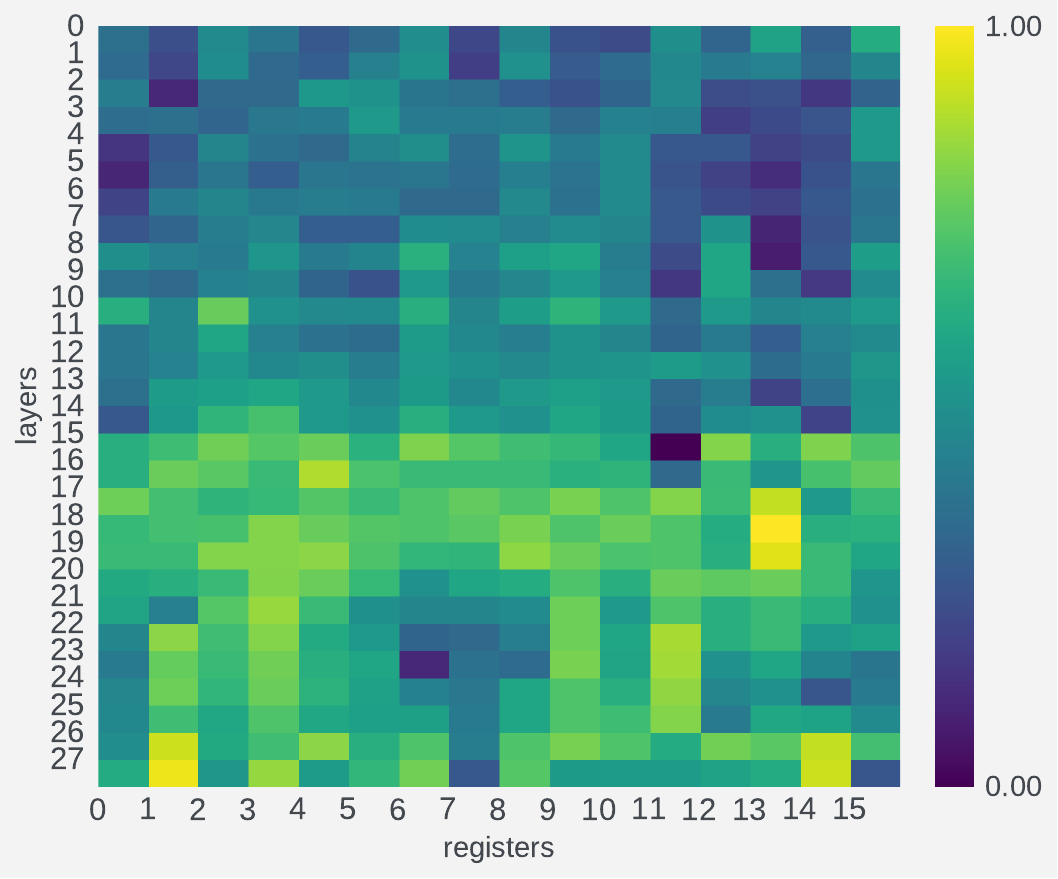}
        \caption{$\boldsymbol{\alpha}$ distribution.}
        \label{fig:alpha_vis}
    \end{subfigure}\hfill
    \begin{subfigure}[b]{0.49\linewidth}
        \centering
        \nextfloat
        \includegraphics[width=\linewidth]{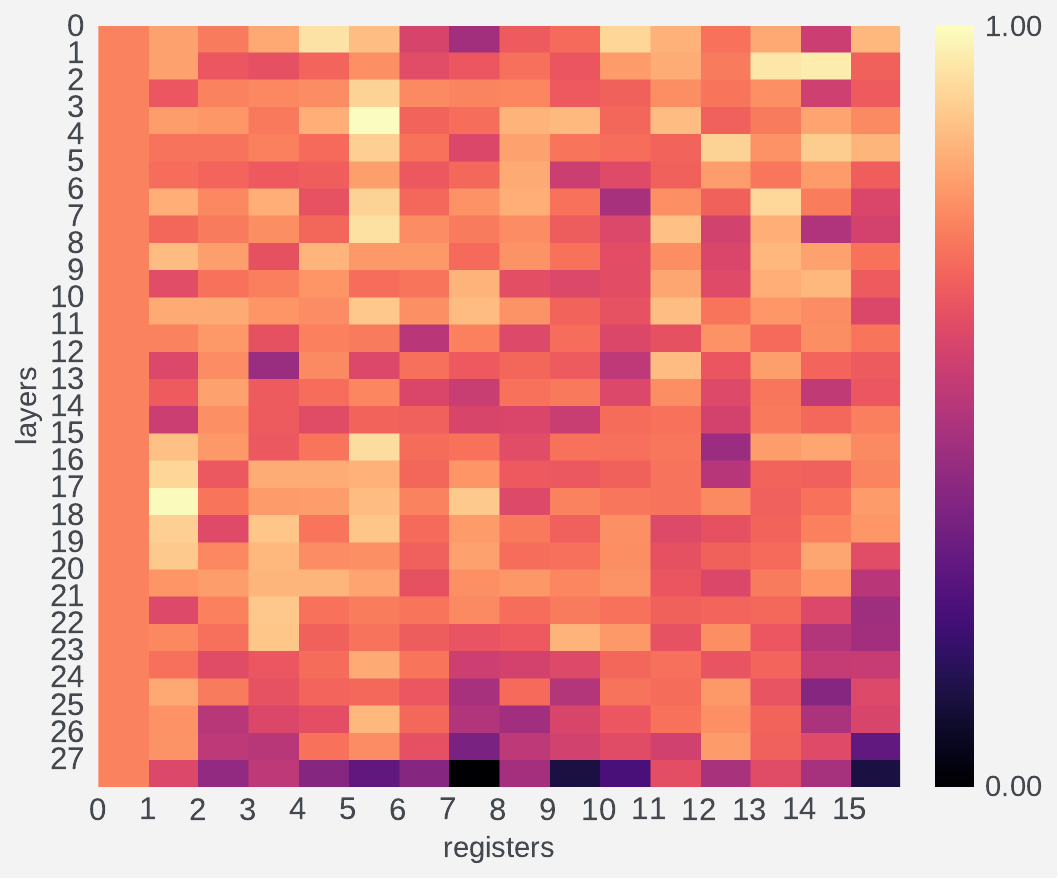}
        \caption{$\boldsymbol{\beta}$ distribution.}
        \label{fig:beta_vis}
    \end{subfigure}
    \caption{Visualization of normalized $\boldsymbol{\alpha}$ and $\boldsymbol{\beta}$ distributions across layers and special token positions when all layers (0-27) are applied with DepThink.}
    \label{fig:alpha_beta_vis}
\end{minipage}
\end{figure}

\subsection{Scaling Model Size and Architecture}
\label{app:scaling_compare}

\textbf{Multimodal retrieval.} In the main paper, the default recipe of DepThink is the supervised fine-tuning the Qwen3-VL-2B-Instruct by LoRA, aka, DepThink-2B-LoRA. Going beyond this setting, we scale up DepThink to additional settings:
\begin{itemize}
    \item \textbf{2B, full FT}: keep using 2B model size, but change the LoRA to full model fine-tuning.
    \item \textbf{8B, LoRA}: scale up the dense model size to 8B backbone Qwen3-VL-8B-Instruct, and keep training with the same LoRA recipe as DepThink-2B-LoRA.
    \item \textbf{30B-A3B MoE, LoRA}: scale up the dense model to the 30B sparse mixture-of-expert (MoE) backbone Qwen3-VL-30B-A3B-Instruct which has only 3B active parameters, and keep training with the same LoRA recipe as DepThink-2B-LoRA.
\end{itemize}
In Tab.~\ref{app:tab:scale_architecture}, we see DepThink is best under all settings comparing with LoopLM and the DepThink without learnable residuals. These results show that the observed gains are not confined to LoRA, the 2B scale, or a dense backbone, although all evaluated backbones remain within the Qwen3-VL family.

\textbf{GSM8K math reasoning.} Similar to the model scaling on multimodal retrieval, we perform the same scaling experiments on GSM8K dataset to validate them for math reasoning. In Tab.~\ref{tab:scaling_gsm8k}, we see that DepThink-8B-LoRA achieved the best performance, even higher than the 30B-A3B MoE model. This is primarily because the dataset GSM8K is relatively small such that the capability of 30B model may not evident on this dataset.

\begin{table}[t]
\centering
\footnotesize
\setlength{\tabcolsep}{4pt}
\begin{threeparttable}[t]
\caption{Model scaling of DepThink on GSM8K.}
\label{tab:scaling_gsm8k}
\begin{tabular}{lc}
\toprule
Setting & Accuracy \\
\midrule
2B, LoRA & 71.11 \\
2B, full fine-tune & 72.20 \\
8B, LoRA & \textbf{85.76} \\
30B-A3B MoE, LoRA & 84.31 \\
\bottomrule
\end{tabular}
\end{threeparttable}
\hfill
\begin{threeparttable}[t]
\caption{Layer-wise extension of DepThink on OK-VQA.}
\label{tab:layer_iid_ablation}
\begin{tabular}{lcc}
\toprule
Residual Layers & PPL & CE \\
\midrule
Early (0--7) & 2.0455 & 0.6810 \\
Middle (8--19) & \textbf{2.0175} & \textbf{0.6677} \\
Late (20--27) & 2.1591 & 0.7336 \\
All (0--27) & 2.0483 & 0.6814 \\
\bottomrule
\end{tabular}
\end{threeparttable}
\hfill
\begin{threeparttable}[t]
\caption{Wall time (s/step, train\,/\,inference) on 16 H100 GPUs.}
\label{tab:latency}
\begin{tabular}{cccc}
\toprule
$K$ & LoopLM & DepThink & (-residual) \\
\midrule
2  & 3.13\,/\,2.01  & 1.74\,/\,1.69 & 1.74\,/\,1.66 \\
4  & 5.77\,/\,2.73  & 1.91\,/\,1.71 & 1.80\,/\,1.80 \\
8  & 11.25\,/\,4.20 & 1.81\,/\,1.68 & 1.75\,/\,1.69 \\
16 & 22.04\,/\,7.24 & 1.85\,/\,1.70 & 1.78\,/\,1.67 \\
\bottomrule
\end{tabular}
\end{threeparttable}
\end{table}

\subsection{Easy vs Hard retrieval tasks.} 
To test DepThink's benefit on task difficulty, we compare {DepThink} and its ablation without residual on video retrieval ({v\_ret}), which requires fine-grained alignment between spatio-temporal video content and language, and zero-shot image classification ({i\_cls}), a much easier task.
As shown in Figure~\ref{fig:video_easy_hard}, {DepThink} outperforms {its ablation} at every register count, with an average relative gain of {2.33\%} on {v\_ret}, nearly six times the {0.41\%} on {i\_cls}.
This suggests that DepThink helps most on harder tasks: when visual-textual alignment is already easy, depth-linked residuals add little, but on {v\_ret} the extra capacity helps capture the fine-grained semantics that retrieval needs.

\subsection{Efficiency Benchmark}
\label{app:efficiency}

\textbf{Experimental details.} To benchmark the training and inference efficiency, we used 16 H100 GPUs, FSDP2, and bf16 precision for each model. The per-GPU batch size is 32, the maximum sequence length is 2048, and a fixed seed provides the same batch stream. Training latency is averaged over 150 steps after 50 warm-up steps. Inference latency is averaged over 200 evaluation steps. 

\textbf{Efficiency results.} In Tab.~\ref{tab:latency}, it clearly shows that either the learnable residuals from DepThink, or the increasing number of special tokens ($K$), has almost no impact on the average wall-time in both training and inference. Noticetably, the LoopM wall-time shows an significant increasing trend when using larger number of looping, especially in training. These impirical findings are consistent with our theoretical analysis of Section~\ref{sec:efficient_inference} in the main paper.

\subsection{Layer-wise Extension.} Thanks to the generality of DepThink, we empirically extend DepThink by applying it to multiple Transformer layers. We evaluate the text generation performance on LLaVA-CoT dataset by perplexity (PPL) and cross-entropy (CE) in Tab.~\ref{tab:layer_iid_ablation}. 
It shows applying DepThink only to the middle layers performs best, even better than applying it to all layers. It also shows applying DepThink only to the late layers performs worst, suggesting DepThink is useful for early-to-middle feature abstraction when extending to multiple layers. 
In Fig.~\ref{fig:alpha_beta_vis}, we visualize the distributions of $\boldsymbol{\alpha}$ and $\boldsymbol{\beta}$ across layers and special tokens for the all-layers version. The results clearly show high $\boldsymbol{\alpha}$ activation from the middle to late layers. This resonates with the findings in~\citep{song2026demystifying} which demonstrates that these layers are critical for reasoning tasks. Moreover, the low values of $\boldsymbol{\beta}$ in the last few layers indicate a saturated contribution from historical latent thoughts in deeper layers.

% % \clearpage
% \input{tables/demo_table}

% \section{Text generation examples}
% \label{sec:text-generation-examples}
% \label{tab:qual_vqa_2_main}

% \clearpage

% \section{Text generation examples}
% \label{sec:text-generation-examples}

% \onecolumn
% \input{tables/demo_table}

\clearpage
% \onecolumn
% \FloatBarrier
\section{Multimodal Reasoning Examples}
\label{sec:text-generation-examples}
% ============================================================
% Reformatted: 10 single-column tables (one per example),
% each spanning the FULL page (line) width via tabularx.
%
% Requires in preamble:
%   \usepackage{tabularx}
%   \usepackage{booktabs}
%   \usepackage{graphicx}
%   \usepackage{xcolor}
%   \usepackage{adjustbox}
% Assumes the following macros/colors are defined elsewhere:
%   \qualtitle, \qualcell, \tokbadge, \rightans, \wrongans, bodytext color.
% (If \qualcell is hard-coded to a fixed width, the image+question row
%  will still render fine; only the reasoning rows need the full width.)
% ============================================================

% ============================================================
% Table 1
% ============================================================
\begin{table}[!htbp]
\centering\small\sffamily\color{bodytext}
\renewcommand{\arraystretch}{1.1}

\begin{adjustbox}{max totalheight=0.85\textheight}
\begin{tabularx}{\linewidth}{@{}X@{}}
\toprule
\qualtitle{} \\
\midrule
\qualcell{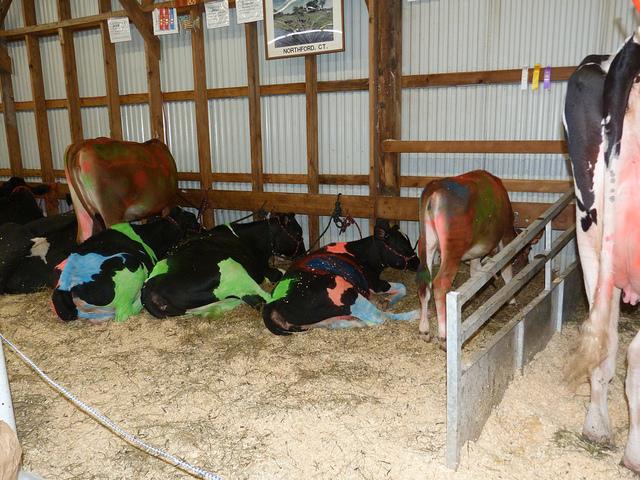}
  {What kind of building are these animals in?}{barn}
\\
\midrule
\tokbadge{2}\,I will analyze the visual characteristics of the barn and the animals to determine the type of building. The image shows a large, open-sided structure with a wooden frame and a roof made of wood planks. The interior is filled with hay bales and several cows. The walls are constructed from wooden beams and panels, and the roof is also made of wood. The building's construction from wood and its open design are typical features of a barn. Bovines are commonly kept in barns for their milk and meat production. \rightans{barn}
\\[4pt]
\tokbadge{4}\,I will identify the type of building shown in the image by observing its structure and features. The image shows a large, enclosed space with wooden walls and a roof made of corrugated metal sheets. There are several animals, including cows and sheep, standing inside. The floor is covered with hay or straw, and there are wooden stalls or pens visible. The presence of animals like cows and sheep, along with the wooden walls and corrugated metal roof, is characteristic of a barn or stable. These structures are specifically designed for housing livestock. \rightans{barn}
\\[4pt]
\tokbadge{8}\,I will analyze the visual evidence in the image to determine the type of building where the animals are kept. The image shows several cows and calves standing in a large, open-sided structure with wooden beams and a roof. The floor is covered with straw or hay, and there are visible signs of wear and tear on the walls. The presence of animals like cows and calves in a large, open-sided building with a roof and straw flooring is characteristic of a barn or stable. These structures are specifically designed for housing livestock. \rightans{barn}
\\[4pt]
\tokbadge{16}\,I will identify the type of building by observing the architectural features and the animals present. The image shows several cows with colorful markings inside a structure with wooden beams and a sloped roof. The walls appear to be made of wood or a similar material. The presence of wooden beams, a sloped roof, and the overall construction style are characteristic of a barn. Barns are specifically designed for housing livestock such as cows. \rightans{barn}
\\
\bottomrule
\end{tabularx}
\end{adjustbox}
\caption{OK-VQA qualitative results across register-token counts.
Final predictions are marked
\rightans{correct} or \wrongans{wrong}.}
\label{tab:qual_vqa_1}
\end{table}

% ============================================================
% Table 2
% ============================================================
\begin{table}[ht]
\centering\footnotesize\sffamily\color{bodytext}
\renewcommand{\arraystretch}{1.1}
\begin{adjustbox}{max totalheight=0.85\textheight}
\begin{tabularx}{\linewidth}{@{}X@{}}
\toprule
\qualtitle{} \\
\midrule
\qualcell{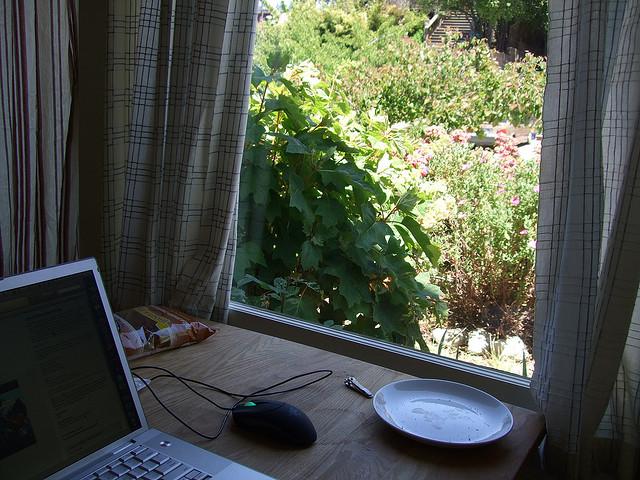}
  {What object is synonymous with a rodent in this image?}{mouse}
\\
\midrule
\tokbadge{2}\,I will identify the object in the image that matches the description of a ``cucumber'' and then determine its function. The image shows a wooden table with a laptop, a glass of water, and a small green object on the left side. The green object has a long, slender shape and a distinct handle, which is characteristic of a cucumber. A cucumber is a common vegetable, and it is typically used as a garnish or condiment in various dishes. \wrongans{cucumber}
\\[4pt]
\tokbadge{4}\,I will identify the rodent in the image and determine what object it is interacting with. A silver laptop is open on a wooden desk. To the left of the laptop, there is a silver rodent, likely a mouse, with its head and front paws visible. The rodent is positioned directly in front of the laptop, which is the primary object of interest in the scene. \wrongans{laptop}
\\[4pt]
\tokbadge{8}\,I will identify the rodent in the image and determine what object it is interacting with. The image shows a cluttered desk with a laptop, a mouse, a mug, and a plate of food. A small rodent is visible on the left side of the desk. The rodent is positioned directly on the surface of the laptop, which is the primary object it is interacting with. \wrongans{laptop}
\\[4pt]
\tokbadge{16}\,I will identify the rodent-like object in the image and find its corresponding synonym in the provided options. A person is sitting at a table with a laptop, a plate of food, and a window with blinds. A small rodent, likely a mouse, is visible on the table. The rodent-like object is a mouse. The word ``mouse'' is one of the options provided. \rightans{mouse}
\\
\bottomrule
\end{tabularx}
\end{adjustbox}
\caption{OK-VQA qualitative results across register-token counts.}
\label{tab:qual_vqa_2}
\end{table}

% ============================================================
% Table 3
% ============================================================
\begin{table}[ht]
\centering\footnotesize\sffamily\color{bodytext}
\renewcommand{\arraystretch}{1.1}
\begin{adjustbox}{max totalheight=0.85\textheight}
\begin{tabularx}{\linewidth}{@{}X@{}}
\toprule
\qualtitle{} \\
\midrule
\qualcell{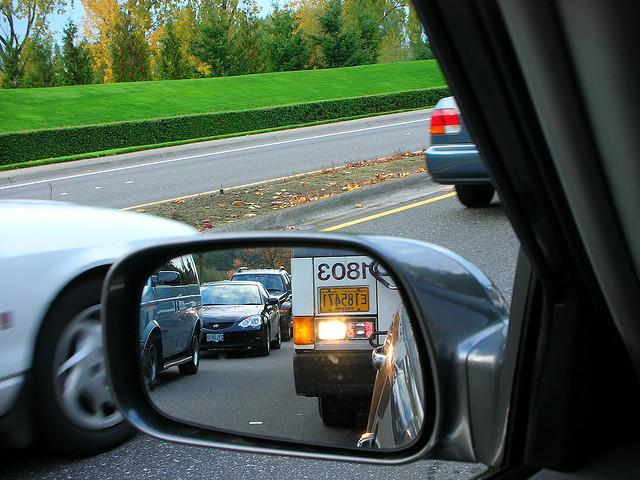}
  {What season is it most likely in the image?}{fall}
\\
\midrule
\tokbadge{2}\,I will analyze the visual cues in the image to determine the season. The image shows a car driving on a road with trees lining the sides. The trees have green leaves, and the sky is overcast with clouds. The overall scene appears to be in a temperate climate. The presence of green leaves on the trees indicates that the season is likely spring or summer. The overcast sky and the presence of green foliage are characteristic of these seasons. \wrongans{Spring}
\\[4pt]
\tokbadge{4}\,I will analyze the visual cues in the image to determine the season. The image shows a street scene with a car, trees, and a road sign. The trees have leaves, and the sky is overcast. The presence of leaves on the trees indicates that the season is not winter. The overall lighting and the appearance of the foliage suggest it is not summer. Therefore, the most likely season is autumn. \rightans{autumn}
\\[4pt]
\tokbadge{8}\,I will analyze the visual cues in the image to determine the season. The image shows a view from inside a car, looking out at a road with trees and a distant building. The trees have leaves, but they appear somewhat sparse and yellowish, suggesting they are not in their peak green state. This indicates that the season is likely not summer. summer The image shows a view from inside a car, looking out at a road with trees and a distant building. The trees have leaves, but they appear somewhat sparse and yellowish, suggesting they are not in their peak green state. This indicates that the season is likely not summer. \rightans{fall}
\\[4pt]
\tokbadge{16}\,I will analyze the visual cues in the image to determine the season. The image shows a car on a road with trees lining the sides. The trees have green leaves, but some appear to be turning yellow or brown, which is typical for autumn. The sky is overcast, and the overall lighting suggests it might be early morning or late afternoon. The presence of changing foliage colors from green to yellow and brown indicates that the season is transitioning from summer to fall. \rightans{autumn}
\\
\bottomrule
\end{tabularx}
\end{adjustbox}
\caption{OK-VQA qualitative results across register-token counts.}
\label{tab:qual_vqa_3}
\end{table}

% ============================================================
% Table 5
% ============================================================
\begin{table}[ht]
\centering\small\sffamily\color{bodytext}
\renewcommand{\arraystretch}{1.25}
\begin{adjustbox}{max totalheight=0.85\textheight}
\begin{tabularx}{\linewidth}{@{}X@{}}
\toprule
\qualtitle{} \\
\midrule
\qualcell{}
  {What type of ship is this?}{cruise}
\\
\midrule
\tokbadge{2}\,I will analyze the visual characteristics of the ship to determine its type. The image shows a large, multi-decked vessel docked at a pier. It has a prominent white superstructure with a red roof, multiple windows, and a large, open-air deck. The ship is moored to a concrete pier with a railing. The vessel's design, featuring a large, open-air deck and a multi-deck structure, is characteristic of a cruise ship. Cruise ships are designed for passenger travel and typically have large, open decks for sightseeing and dining. \rightans{cruise ship}
\\[4pt]
\tokbadge{4}\,I will identify the type of vessel based on its visible features and size. The image shows a large, white passenger ship docked at a port. It has multiple decks, numerous windows, and a prominent funnel. The ship is connected to a pier by a gangway. The vessel's size, multiple decks, and numerous windows are characteristic of a passenger liner or cruise ship, which is designed to carry a large number of people for leisure travel. \rightans{cruise ship}
\\[4pt]
\tokbadge{8}\,I will analyze the visual characteristics of the large vessel to determine its type. The image shows a very large, white passenger ship docked at a pier. It has multiple decks, numerous windows, and a prominent bridge structure at the front. The vessel's size, number of decks, and overall design are characteristic of a cruise ship or a large passenger liner. These features are typical for ships designed to carry a significant number of passengers for leisure or tourism purposes. \rightans{cruise ship}
\\[4pt]
\tokbadge{16}\,I will identify the type of boat based on its size, design, and features visible in the image. The image shows a large, multi-deck vessel with a prominent bow, a large central structure, and multiple decks. It is situated on a body of water with other boats and a city skyline in the background. The boat's size, multiple decks, and overall design are characteristic of a cruise ship. Cruise ships are designed for passenger comfort and are typically large enough to accommodate hundreds of people. \rightans{cruise ship}
\\
\bottomrule
\end{tabularx}
\end{adjustbox}
\caption{OK-VQA qualitative results across register-token counts.}
\label{tab:qual_vqa_5}
\end{table}

% ============================================================
% Table 7
% ============================================================
\begin{table}[ht]
\centering\small\sffamily\color{bodytext}
\renewcommand{\arraystretch}{1.25}
\begin{adjustbox}{max totalheight=0.85\textheight}
\begin{tabularx}{\linewidth}{@{}X@{}}
\toprule
\qualtitle{} \\
\midrule
\qualcell{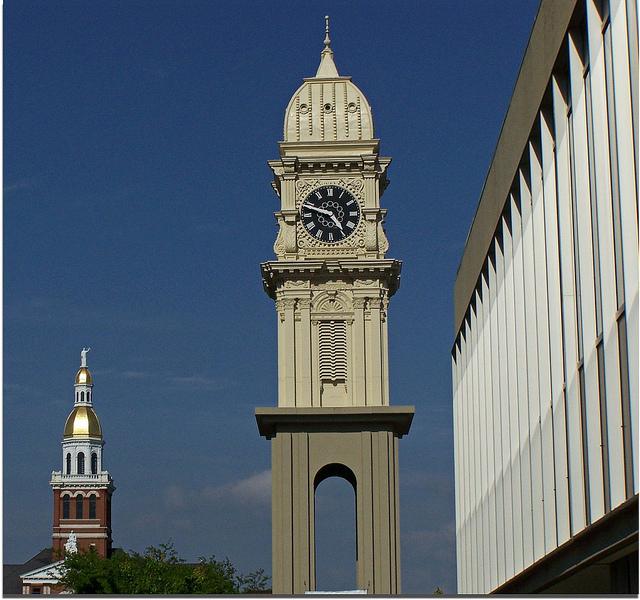}
  {Are the tops of these towers minarets or domes?}{minaret}
\\
\midrule
\tokbadge{2}\,I will identify the architectural style of the dome and compare it to the other structures shown. The image displays a tall, slender structure with a large, domed roof. The dome is made of a light-colored material, possibly stone or concrete, and has a distinct curved shape. The structure is supported by a series of vertical columns. The dome's design, with its large, curved roof and the way it is supported by columns, is characteristic of a dome. This architectural style is commonly used in churches, mosques, and other religious buildings. \wrongans{dome}
\\[4pt]
\tokbadge{4}\,I will analyze the architectural style of the tower to determine if it is a clock tower or a minaret. The image shows a tall, slender structure with a domed roof and a small, pointed spire at the top. The base of the structure is a large, circular platform. The design features a dome and a spire, which are characteristic elements of a minaret. Clock towers typically have a more rectangular or square base and a clock face, not a dome. \rightans{minaret}
\\[4pt]
\tokbadge{8}\,I will analyze the architectural features of the two structures to determine which one is a minaret. The image shows two tall structures against a clear blue sky. The one on the left has a dome and a spire, while the one on the right has a tall, narrow tower with a conical roof. The structure on the right is a minaret, which is a tall, slender tower used for the call to prayer. The structure on the left is a dome, which is a type of roof or covering for a building. Therefore, the structure on the right is a minaret. \rightans{minaret}
\\[4pt]
\tokbadge{16}\,I will identify the architectural style of the tower in the background and compare it to the top of the tower in the foreground to determine if they are the same. The image shows a tall, white tower with a dome on top. In the background, there is another tower with a pointed roof. The tower in the foreground has a distinct pointed roof, which is characteristic of a minaret. The tower in the background also has a pointed roof, which is also characteristic of a minaret. Therefore, both towers are minarets. \rightans{minarets}
\\
\bottomrule
\end{tabularx}
\end{adjustbox}
\caption{OK-VQA qualitative results across register-token counts.}
\label{tab:qual_vqa_7}
\end{table}

% ============================================================
% Table 8
% ============================================================
\begin{table}[ht]
\centering\small\sffamily\color{bodytext}
\renewcommand{\arraystretch}{1.25}
\begin{adjustbox}{max totalheight=0.85\textheight}
\begin{tabularx}{\linewidth}{@{}X@{}}
\toprule
\qualtitle{} \\
\midrule
\qualcell{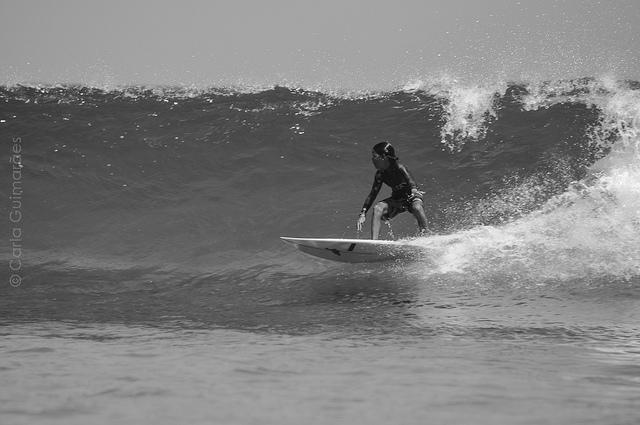}
  {Can you guess the sea name shown in this picture?}{pacific ocean}
\\
\midrule
\tokbadge{2}\,I will identify the type of watercraft shown in the image and match it to the correct category of surfing. The image displays a person riding a large, white, surfboard-like board on a wave. The board has a long tail and appears to be designed for riding waves, which is characteristic of a surfboard. The object in the image is clearly a surfboard, which is a type of watercraft used for surfing. \wrongans{surfboard}
\\[4pt]
\tokbadge{4}\,I will identify the type of wave based on its characteristics and then determine the sea name associated with it. The image shows a large, dark wave breaking over a white surfboard. The wave has a distinct shape with a steep face and a hollowed-out center, which is characteristic of a specific type of wave. The wave's appearance, with its steep face and hollowed-out center, is typical of a breaking wave. This type of wave is commonly found in the Pacific Ocean, particularly in the region known as the ``Sea of the South Pacific'' or ``South Pacific Sea.'' \wrongans{South Pacific Sea}
\\[4pt]
\tokbadge{8}\,I will analyze the visual elements in the image to identify the location of the surfing activity. The image shows a person riding a large wave on a surfboard. The background features a rocky shoreline and the ocean. The person is wearing a wetsuit and appears to be in motion. The presence of a rocky coastline and the specific style of wave breaking are characteristic of a particular type of surf spot. This particular location is known for its consistent waves and is a popular destination for professional surfers. The surf spot is known for its consistent waves. The presence of a rocky coastline and the specific style of wave breaking are characteristic of a particular type of surf spot. This particular location is known for its consistent waves and is a popular destination for professional surfers. \wrongans{(no answer)}
\\[4pt]
\tokbadge{16}\,I will identify the body of water in the image to determine the sea name. The image shows a person surfing on a large wave in the ocean. The water appears to be a deep blue color, typical of the Pacific Ocean. The visual evidence of the ocean's appearance, specifically its deep blue color and the presence of a large wave, strongly suggests that the location is the Pacific Ocean. \rightans{Pacific Ocean}
\\
\bottomrule
\end{tabularx}
\end{adjustbox}
\caption{OK-VQA qualitative results across register-token counts.}
\label{tab:qual_vqa_8}
\end{table}

% ============================================================
% Table 10
% ============================================================
\begin{table}[ht]
\centering\small\sffamily\color{bodytext}
\renewcommand{\arraystretch}{1.25}
\begin{adjustbox}{max totalheight=0.85\textheight}
\begin{tabularx}{\linewidth}{@{}X@{}}
\toprule
\qualtitle{} \\
\midrule
\qualcell{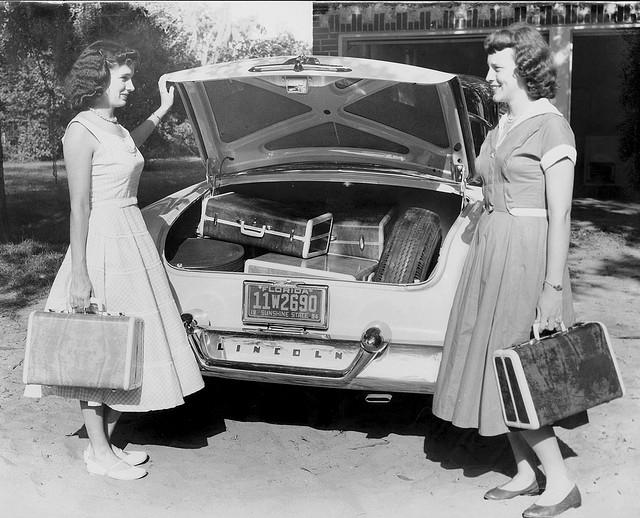}
  {What brand of vehicle is that?}{lincoln}
\\
\midrule
\tokbadge{2}\,I will identify the car model by examining the front of the vehicle and the text on the front bumper. The image shows a vintage car parked in a parking lot. The front of the car has a white bumper with text on it. The text on the bumper reads ``1967'' and ``FORD''. This indicates the year of manufacture and the manufacturer of the vehicle. \wrongans{1967 FORD}
\\[4pt]
\tokbadge{4}\,I will identify the brand of the car by examining its front grille and emblem. The image shows a vintage car with a large, prominent emblem on the front grille. The text ``LINCOLN'' is clearly visible above the emblem. The emblem on the front of the vehicle is the Lincoln logo, which is a well-known brand of automobiles. \rightans{lincoln}
\\[4pt]
\tokbadge{8}\,I will identify the brand of the car by examining its distinctive design elements, such as the shape of the headlights and the overall body style. The image shows a vintage car with a prominent hood ornament, a large trunk, and a distinctive front grille. The car's design features a rounded hood and a prominent chrome grille with a central emblem. The car's design, particularly the rounded hood and the prominent chrome grille with a central emblem, is characteristic of a specific car brand. This design is famously associated with the Lincoln Continental. \rightans{Lincoln}
\\[4pt]
\tokbadge{16}\,I will identify the brand of the vehicle by examining the logo on the front of the car. The image shows a vintage car with a trunk open, revealing luggage. The car has a distinct emblem on the front grille. The emblem on the front of the car is clearly the Lincoln Motor Company logo, which is a stylized ``L'' within a circle. \rightans{lincoln}
\\
\bottomrule
\end{tabularx}
\end{adjustbox}
\caption{OK-VQA qualitative results across register-token counts.}
\label{tab:qual_vqa_10}
\end{table}

\end{document}